\documentclass[11pt]{article}

\usepackage[final]{acl}

\usepackage{times}
\usepackage{latexsym}
\usepackage{float}
\usepackage{placeins}
\usepackage{booktabs}
\usepackage[table]{xcolor}
\definecolor{soberblue}{RGB}{70,95,135}
\usepackage{multirow}
\usepackage{CJKutf8}
\usepackage{amsmath}
\usepackage{amssymb}
\usepackage{caption}
\usepackage{enumitem}
\usepackage[T1]{fontenc}

\usepackage[utf8]{inputenc}

\usepackage{microtype}

\usepackage{inconsolata}

\usepackage{graphicx}
\usepackage{fvextra}
\usepackage{paracol}

\title{Rethinking the Multilingual Reasoning Gap with Layer Swap}

\author{
\textbf{Maxence Lasbordes}$^{1,2}$ \quad
Amélie Chatelain$^{1}$ \quad
Djamé Seddah$^{2}$ \\
$^{1}$LightOn, Paris \quad
$^{2}$Inria, Paris \\
\texttt{\{maxence.lasbordes, amelie\}@lighton.ai} \\
\texttt{djame.seddah@inria.fr}
}

\begin{document}
\maketitle
\begin{abstract}
Recent reasoning Large Language Models produce a chain-of-thought (CoT) predominantly in English, even when prompted in non-English languages. Prior work suggests that forcing the CoT to remain in the input language (\emph{native reasoning}) substantially degrades performance relative to allowing the model to reason in English before answering in the input language (\emph{English-pivoted reasoning}). However, most studies of this native reasoning gap rely on inference-time interventions or limited native-language training data. We revisit this comparison at a larger scale and under comparable supervision. We construct long multilingual reasoning datasets across six languages (English, French, German, Spanish, Chinese and Swahili); fine-tune specialists in both native and English-pivoted regimes on top of \texttt{Qwen/Qwen3-8B-Base}, and evaluate across mathematics, science, general knowledge, and code. In this setting, the average native reasoning gap shrinks to 1.9--3.5\% across the five non-English languages, considerably smaller than previously reported. Weight-space analysis of the native specialists reveals aligned fine-tuning updates in the middle layers and divergence in the outer layers. This points to a largely language-agnostic reasoning core surrounded by language-specific layers. Exploiting this structure, we introduce a Layer Swap: transferring the English specialist's stronger reasoning mid-layers into each native specialist, closing most of the native reasoning gap across the five non-English languages while preserving CoT in the target language. We release all models and datasets.\footnote{\url{https://huggingface.co/collections/lightonai/multilingual-reasoning}}
\end{abstract}

\begin{figure*}[t]
    \centering
    \includegraphics[width=\textwidth]{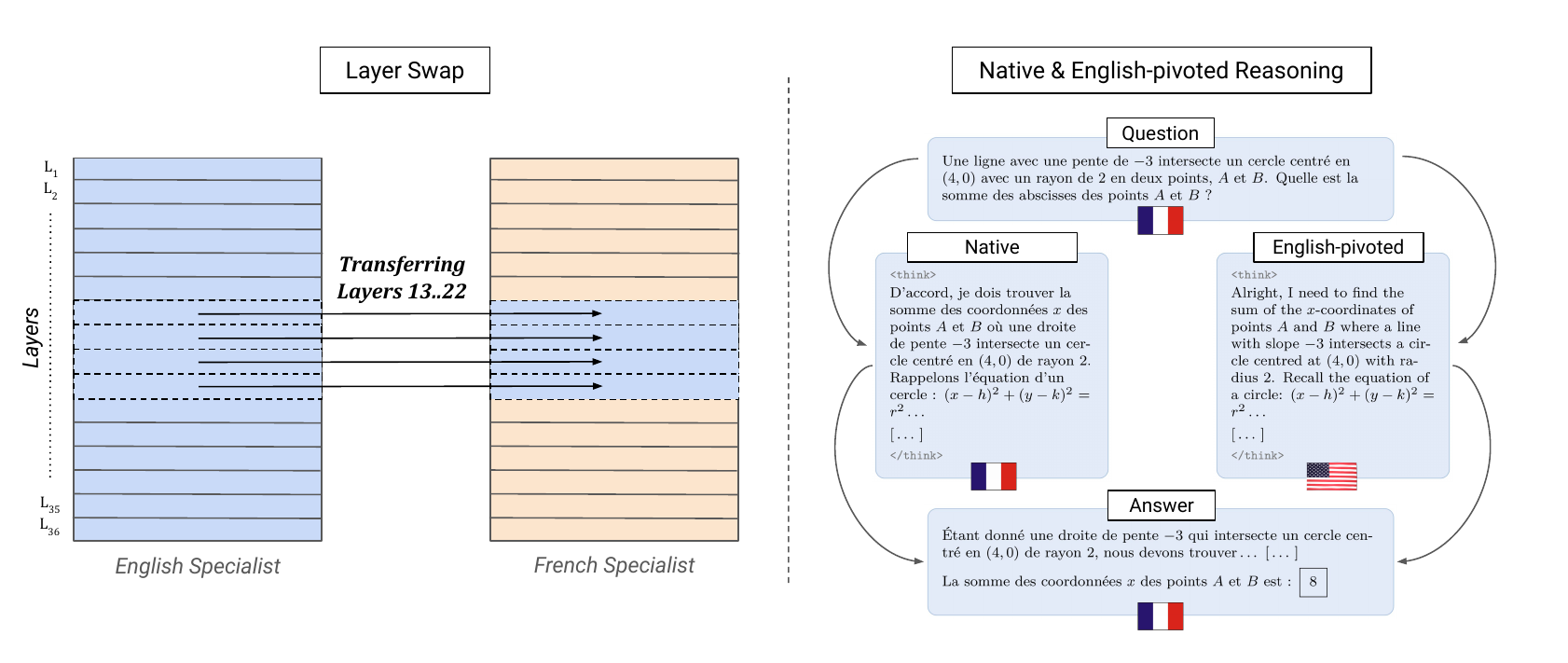}
    \caption{(\emph{left}) \textbf{Layer Swap}: transferring a mid-stack window of the English specialist into the native specialist keeps the CoT in the input language while reducing the remaining native reasoning gap. (\emph{right}) The two baselines compared in this work: \textbf{Native Reasoning} (CoT in the input language, here French) and \textbf{English-pivoted Reasoning} (CoT in English regardless of input), the default of open multilingual reasoning models.}
    \label{fig:native-vs-pivoted}
\end{figure*}

\section{Introduction}
Reasoning models such as OpenAI o1~\citep{jaech2024openai}, DeepSeek-R1~\citep{guo2025deepseek}, and Qwen3~\citep{yang2025qwen3} rely on long chain-of-thought (CoT) to tackle complex tasks in mathematics, code, and science. In current open reasoning models, the CoT is overwhelmingly produced in English, including on non-English inputs: the model switches back to the input language only for the final response~\citep{saji2026reasoning, park2025cross}, a regime we refer to as \emph{English-pivoted reasoning} (Figure~\ref{fig:native-vs-pivoted}). This default has practical costs. English-only reasoning limits interpretability for non-English users, reduces the linguistic and cultural nuance captured by native reasoning traces, and accumulates translation-style errors that compound with task complexity~\citep{saji2026reasoning}. Constraining the CoT to remain in the input language, \emph{native reasoning} (Figure~\ref{fig:native-vs-pivoted}), is therefore desirable, but most attempts substantially degrade accuracy on key benchmarks~\citep{barua2025long, zhang2025think}. We call models trained for native reasoning in a specific non-English language \emph{native specialists}. However, these measurements are typically obtained either through prompting alone~\citep{saji2026reasoning, kang2025multilingual} or through fine-tuning on small amounts of native reasoning data~\citep{barua2025long, qi2025modelsreasonlanguagecontrolling}, using post-training budgets far below those required for meaningful native reasoning supervision. Whether the reported gap persists once native post-training is brought to a scale comparable to English-pivoted post-training remains, to our knowledge, understudied.

\begin{figure*}[t]
    \centering
    \includegraphics[width=0.95\textwidth]{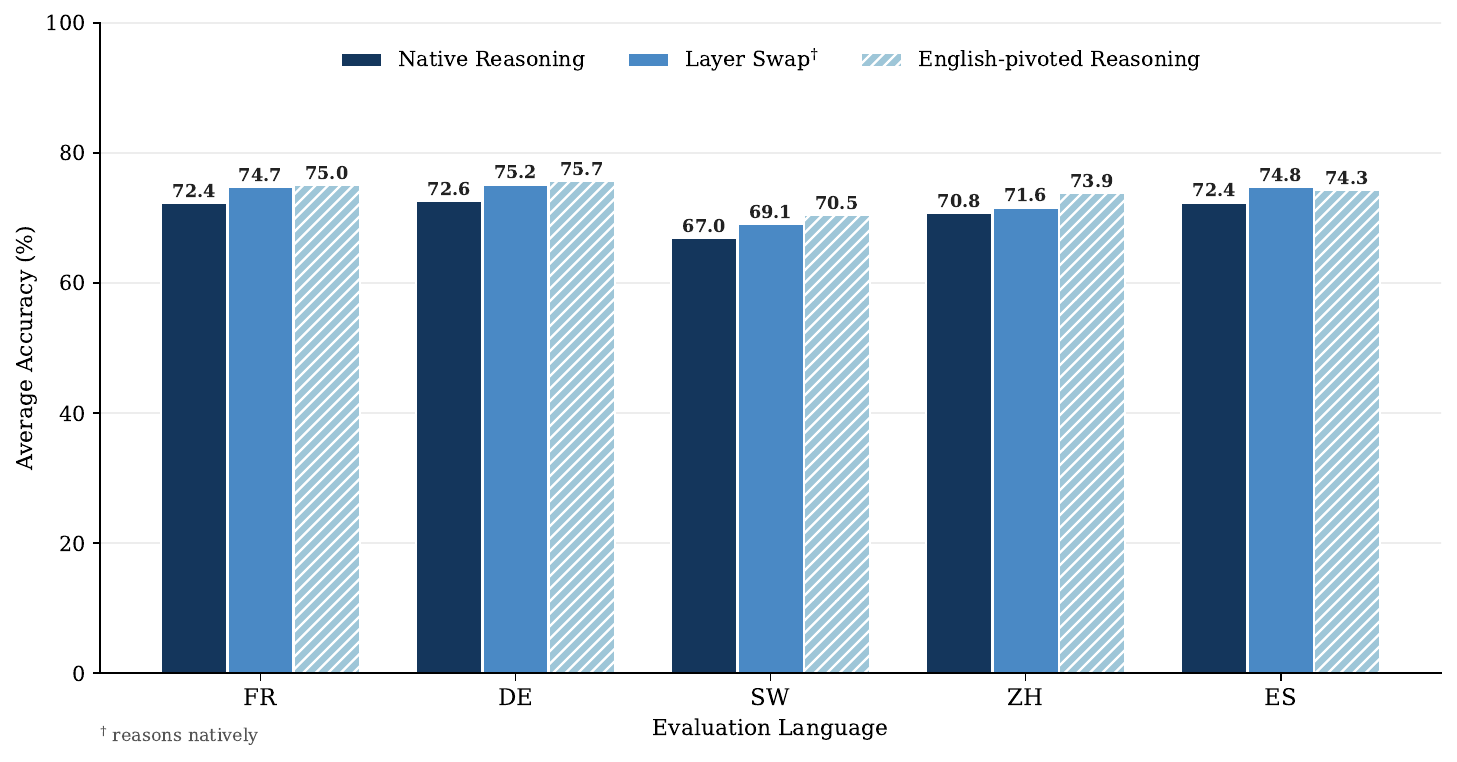}
    \caption{
        Average accuracy across MGSM-Rev2, Global-MMLU-Lite, GPQA-Diamond, AIME 24/25, and HumanEvalPlus in the target language, for \textsc{xx} $\in$ \{\textsc{fr}, \textsc{de}, \textsc{es}, \textsc{zh}, \textsc{sw}\}. Three settings are compared per language: \textbf{Native Reasoning} (\texttt{Qwen3-8B-xx}, CoT in \textsc{xx}), \textbf{Layer Swap} (\texttt{Qwen3-8B-xx-Swap}, a mid-stack window of the English specialist transferred into the native specialist, CoT in \textsc{xx}; see \S\ref{sec:layer-swap}), and \textbf{English-Pivoted Reasoning} (\texttt{Qwen3-8B-xx-Pivot-EN}, CoT in English).
    }
    \label{fig:main_figure_comparison}
\end{figure*}

In this work, we revisit the comparison under strictly comparable training conditions. We construct a large multilingual reasoning dataset spanning six languages (French, German, Spanish, Swahili, Chinese, and English), with approximately 500k samples per language up to 32k tokens, and perform supervised fine-tuning (SFT) on \texttt{Qwen/Qwen3-8B-Base} with roughly 10B tokens per language in both regimes (native and English-pivoted). Across general knowledge, mathematics, code and science benchmarks, the native reasoning gap shrinks to 1.9--3.5\% on average across the five non-English languages (Figure~\ref{fig:main_figure_comparison}), substantially smaller than previous reports suggest~\citep{barua2025long, qi2025modelsreasonlanguagecontrolling}; the gap concentrates on complex reasoning benchmarks, with much smaller gaps on the other benchmarks. Weight-space analysis of the per-language SFT updates further shows that cross-language updates align tightly in the middle layers but diverge at the edges. We exploit this structure with a \emph{Layer Swap}~\citep{bandarkar2024layer} (Figure~\ref{fig:native-vs-pivoted}), a training-free method that transfers the English specialist's middle layers into each native specialist. To our knowledge, this technique has not previously been applied to long-CoT reasoning models, nor to pairs of experts that share the same reasoning skills but differ in their training language, the setting we study here. Layer Swap closes $83$--$89$\% of the gap on French and German, $60$\% on Swahili, $27$\% on Chinese, and matches the English-pivoted ceiling on Spanish, all while keeping the CoT in the target language (Figure~\ref{fig:main_figure_comparison}).

Our contributions are: \textbf{(i)} a publicly released long-CoT reasoning corpus across six languages at 32k context, with CoTs in the target language, covering both European and non-European languages, including Swahili and Chinese; \textbf{(ii)} a large-scale, strictly controlled measurement of the native-vs-English-pivoted reasoning gap under matched SFT token budgets; \textbf{(iii)} a weight-space analysis revealing a largely language-agnostic reasoning core in the middle layers, and a training-free Layer Swap that exploits this structure to close most of the remaining gap while preserving the target-language CoT; and \textbf{(iv)} an input-language ablation showing that input understanding remains a primary bottleneck under matched native SFT.

\section{Related Work}

\paragraph{Native multilingual reasoning training}
Recent work has approached the native reasoning gap through both data and post-training objectives with various degrees of success. Publicly available native long-CoT corpora remain scarce~\citep{ghosh2025survey}: existing releases provide only a few hundred to a few thousand samples per language~\citep{barua2025long, qi2025modelsreasonlanguagecontrolling}, while broader multilingual corpora either retain English CoT or do not target long-form reasoning. Within this limited regime, \citet{barua2025long} show that translated long-CoT supervision can effectively train non-English reasoners, motivating our translated-data pipeline at substantially larger scale across six languages.

Post-training approaches similarly reveal a substantial native reasoning gap. Under pure SFT on \texttt{Qwen/Qwen3-8B-Base}, \citet{barua2025long} find that target-language CoT underperforms English-pivoted reasoning, with AIME 24/25 gaps averaging $\sim$19\% across nine languages ($\sim$17\% for French). \citet{son2025pushing} report a similar gap in Korean and mitigate it by inserting English anchor segments into the reasoning trace. Reinforcement learning-based (RL) methods offer mixed evidence: \citet{huang2025beyond} show that RL on non-English data can improve cross-lingual transfer, whereas \textit{Cross-lingual Collapse}~\citep{park2025cross} identifies a recurrent failure mode in which CoT drifts back to English as accuracy improves under GRPO~\citep{shao2024deepseekmath}. Hybrid approaches combine both stages: \textit{Think Natively}~\citep{zhang2025think} applies SFT followed by GRPO with language-consistency and cross-lingual alignment rewards, while concurrent independent work, \textit{ReasonXL}~\citep{gurgurov2026reasonxl}, combines SFT and RLVR on SmolLM3-3B across five European languages at 16k context. Like several prior works, \textit{ReasonXL} evaluates trained native specialists against the original base model, which can make it challenging to isolate the effect of native reasoning from the effect of specialization itself. In this work, we instead compare native and English-pivoted specialists trained under identical conditions, differing only in the language used for CoT reasoning. Our experimental design enables this comparison at broader scope on an 8B model: six languages, including Chinese and Swahili; an extended 32k training context; and, for each language, matched native and English-pivoted specialists trained on similar Q\&A data.

\paragraph{English as a latent reasoning language}
Another line of work argues that multilingual LLMs internally route reasoning through English-aligned representations. Using logit-lens probing on Llama-2, \citet{wendler2024llamas} show that intermediate states traverse an English-aligned region before resolving to the target language; \citet{schut2025multilingual} confirm this with activation steering across several languages, finding stronger transfer from English-derived steering vectors than from native-language ones. The same bias appears behaviourally: \citet{etxaniz2024multilingual} show that explicit self-translation into English can outperform direct non-English inference, while~\citet{saji2026reasoning} argues that English-pivoted reasoning introduces ``Lost in Translation'' errors that compound with task complexity. Closest to our setting, ~\citet{kang2025multilingual} attributes most of the multilingual reasoning gap to input understanding rather than the reasoning process itself. We revisit this decomposition through an additional ablation that varies only the input language across our native specialists, which naturally continue reasoning in their training language without any constraint, isolating the contribution of reasoning language from input understanding.

\section{Data and Benchmarks}

\subsection{Dataset Creation}

Training open-reasoning models in non-English languages requires large native corpora with long-CoT reasoning traces, which remain scarce. We address this gap by constructing such a corpus through automatic translation from an English source, with samples up to $32,768$ tokens across five target languages.

\paragraph{Source corpus}
We start from \emph{allenai/Dolci-Think-SFT-32B}~\citep{olmo2025olmo3}, a decontaminated English post-training dataset covering mathematics, code, instruction following, science, safety, general chat, and structured data. We sample $\sim$500k examples uniformly, preserving the category distribution (Table~\ref{tab:dataset_stats}, Appendix).

\paragraph{Languages}
We target five languages spanning diverse typological and resource settings: French, German, and Spanish (high-resource European, Latin script), Chinese (high-resource, non-alphabetic, typologically distant), and Swahili (low-resource). Together with the original English subset, this yields six per-language corpora of $\sim$500k samples each (Table~\ref{tab:dataset_language_stats}, Appendix).

\paragraph{Translation}
Motivated by prior evidence that translated long-CoT data outperforms direct native distillation~\citep{barua2025long}, we translate the English corpus into the five target languages with \texttt{google/gemma-3-27b-it}, chosen for its strong multilingual coverage. Single-pass translation exhibited two failure modes: on the longest samples (input plus output up to 64K tokens), the model entered a degraded long-context regime with frequent looping; and, independently of length, on a small but consistent fraction of samples it silently dropped the reasoning trace while still translating the question and final answer, a failure that persisted under alternative \verb|<think>| delimiters. We therefore translate each sample component-wise, splitting question, reasoning trace, and final answer into $\sim$2k-token chunks at sentence or paragraph boundaries, translating each chunk independently, and recomposing, at some cost in global coherence. We manually inspected a subset of translations to verify quality and reasoning-trace preservation.

\paragraph{Filtering}
We apply two filtering passes. Before translation, we remove English samples that explicitly reference translating into or answering in a specific language, since they become self-contradictory after translation. After translation, we discard (i) empty outputs, (ii) samples whose zlib~\citep{Gailly2012zlibCL} compression ratio against the source deviates anomalously from the dataset mean, which catches degenerate or repeated outputs, (iii) samples whose translated-to-source length ratio similarly deviates, flagging truncations or over-generation, and (iv) samples whose total length exceeds the 32K-token training context, which disproportionately affects languages with less efficient tokenization (e.g. Swahili) or higher verbosity (e.g. French, Spanish). Chunk-level translation limits per-call context and keeps translation error rates low.

\subsection{Benchmarks}

We evaluate across four different domains:
\begin{itemize}[leftmargin=1em,itemsep=0pt,topsep=2pt]
\item \textbf{Mathematics:} \emph{MGSM-Rev2}~\citep{peter2025mind}, a revised multilingual version of the grade-school mathematics benchmark \emph{GSM8K}~\citep{shi2022language} that corrects translation errors; and multilingual versions of \emph{AIME24} and \emph{AIME25}~\citep{qi2025modelsreasonlanguagecontrolling} for competition-level hard reasoning problems.
\item \textbf{Science:} A multilingual version~\citep{qi2025modelsreasonlanguagecontrolling} of \emph{GPQA-Diamond}~\citep{rein2023gpqa}, a PhD-level science QA benchmark.
\item \textbf{Knowledge:} \emph{Global-MMLU-Lite}~\citep{singh2024globalmmluunderstandingaddressing}, a curated multilingual MMLU benchmark that addresses cultural and translation biases in the original version.
\item \textbf{Code:} We translate \emph{HumanEvalPlus}~\citep{liu2023your} with \texttt{google/gemma-3-27b-it} into our five target languages.
\end{itemize}

\paragraph{Quality control}
The translated reasoning benchmarks (AIME24, AIME25, and GPQA-Diamond)\footnote{\url{https://huggingface.co/collections/lightonai/multilingual-reasoning}} initially contained a small number of translation artifacts that biased evaluation against non-English languages. To mitigate this, we performed an LLM-as-a-judge review using Claude Opus 4.7, which identified and rewrote a few malformed samples in each target language. Evaluation was conducted using a forked version of \texttt{lm-eval-harness}~\citep{eval-harness}; the per-language prompts are listed in Appendix~\ref{sec:eval_prompts}.

\paragraph{Evaluation protocol}
We evaluate models using a temperature of $1$, as lower values caused frequent decoding loops in our models, top-\textit{p} $0.95$, top-\textit{k} $20$, and min-\textit{p} $0$ across all benchmarks. We report mean accuracy over multiple random seeds, using $10$ sampled runs by default, except for AIME24 and AIME25, which we average over $30$ runs to reduce sampling variance on these smaller test sets.

\section{The Native Reasoning Gap Under Matched Supervision}
\subsection{Experimental Setup}
We run distributed training on multiple H100 nodes with TRL~\citep{vonwerra2020trl}. At a 32K sequence length, full fine-tuning of an 8B model exceeds single-H100 memory. To mitigate the issue, we combine DeepSpeed ZeRO-3~\citep{rajbhandari2020zero} parameter sharding with Ulysses sequence parallelism~\citep{jacobs2023deepspeed} that splits attention heads across GPUs, together with FlashAttention-3~\citep{shah2024flashattention} and sequence packing for efficiency. All specialists are fully fine-tuned from \texttt{Qwen/Qwen3-8B-Base} (Table~\ref{tab:sft_key_hparams}).

\subsection{Scaling Across Languages}

\begin{figure}[h]
    \centering
    \includegraphics[width=\columnwidth]{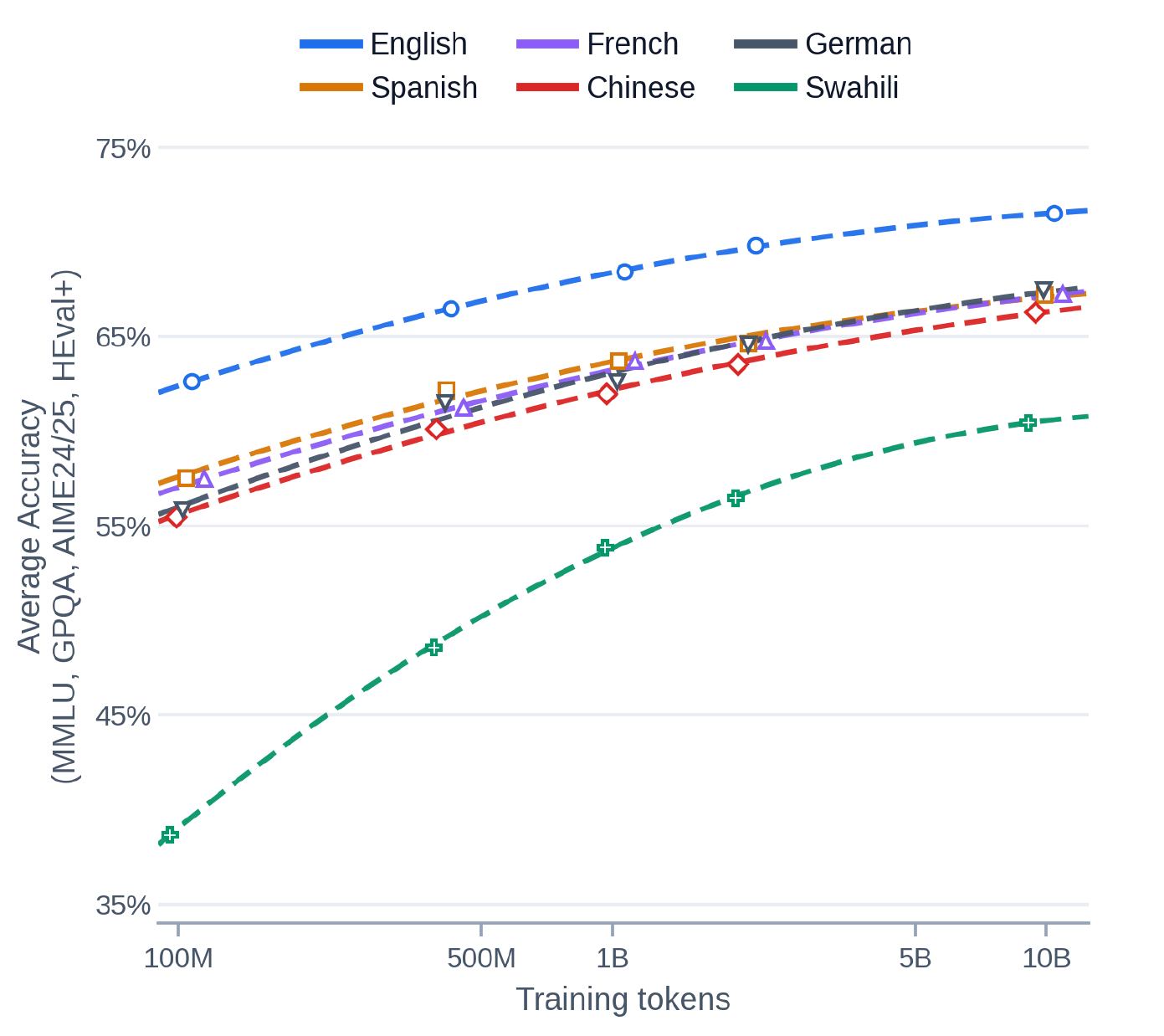}
    \caption{Scaling curves of native reasoning performance across six languages, fine-tuned from \texttt{Qwen/Qwen3-8B-Base}: average accuracy on MGSM-Rev2, Global-MMLU-Lite, GPQA-Diamond, AIME 24/25, and HumanEvalPlus in the training language, as a function of the SFT-token budget.}
    \label{fig:scaling_law_avg}
\end{figure}

Before measuring the native reasoning gap under matched supervision, we verify that our translated corpus produces useful training signal in every language. We fine-tune each of the six per-language corpora at training-data budgets ranging from $\sim$100M to $\sim$10B tokens, holding all other hyperparameters fixed, and evaluate each resulting specialist on our five benchmarks in its training language. Figure~\ref{fig:scaling_law_avg} shows that average accuracy increases monotonically with the budget in every language. This confirms that the translated corpus carries usable training signal in every language. We adopt the $\sim$10B-token budget for the matched-supervision experiment that follows (Table~\ref{tab:training-data-composition}, Appendix).

The curves separate into three resource tiers (Figure~\ref{fig:scaling_law_avg}). English forms the upper envelope at every budget. The high-resource cluster of French, German, Spanish, and Chinese tracks it closely, with a gap that stays in a narrow $4.5$--$6\%$ band across budgets. Swahili sits below this cluster but shows the largest absolute gains, closing its gap to the high-resource tier from roughly $18\%$ at the $\sim$100M-token budget to $7\%$ at $\sim$10B.

\begin{table*}[t]
\centering
\footnotesize
\setlength{\tabcolsep}{2pt}
\renewcommand{\arraystretch}{1.05}
\setlength{\aboverulesep}{0pt}
\setlength{\belowrulesep}{0pt}
\setlength{\extrarowheight}{2pt}
\begin{tabular*}{\textwidth}{@{\extracolsep{\fill}}llccccccc@{}}
\toprule
\textbf{Eval Lang} & \textbf{Model} & \textbf{CoT} & \cellcolor{cyan!15}\textbf{Avg} & \textbf{MGSM} & \textbf{G-MMLU} & \textbf{GPQA-D} & \textbf{AIME 24/25} & \textbf{HEval+} \\
\midrule
\multirow{4}{*}{\textit{French}}  & \texttt{Qwen3-8B-FR}            & {\setlength{\fboxsep}{2pt}\setlength{\fboxrule}{0.7pt}\fcolorbox{soberblue}{white}{\textsc{fr}}} & \cellcolor{cyan!15}$72.36$          & $92.80{\,\scriptstyle\pm1.71}$          & $76.45{\,\scriptstyle\pm1.73}$          & $53.59{\,\scriptstyle\pm2.85}$          & $55.67{\,\scriptstyle\pm3.93}$          & $83.31{\,\scriptstyle\pm1.56}$ \\
                                  & \texttt{Qwen3-8B-FR-Swap}       & {\setlength{\fboxsep}{2pt}\setlength{\fboxrule}{0.7pt}\fcolorbox{soberblue}{white}{\textsc{fr}}} & \cellcolor{cyan!15}$74.74$          & $\mathbf{97.40}{\,\scriptstyle\pm0.60}$ & $76.57{\,\scriptstyle\pm1.54}$          & $54.55{\,\scriptstyle\pm2.74}$          & $59.11{\,\scriptstyle\pm4.08}$          & $\mathbf{86.06}{\,\scriptstyle\pm2.06}$ \\
                                  & \texttt{Qwen3-8B-FR-Pivot-EN}  & \textsc{en} & \cellcolor{cyan!15}$\mathbf{75.04}$ & $94.52{\,\scriptstyle\pm1.28}$          & $\mathbf{78.37}{\,\scriptstyle\pm0.81}$ & $\mathbf{54.65}{\,\scriptstyle\pm2.48}$ & $\mathbf{62.78}{\,\scriptstyle\pm4.84}$ & $84.88{\,\scriptstyle\pm1.53}$ \\
                                  & \texttt{Qwen3-8B-EN}            & \textsc{en} & \cellcolor{cyan!15}$74.27$          & $95.72{\,\scriptstyle\pm0.68}$          & $77.50{\,\scriptstyle\pm1.84}$          & $52.53{\,\scriptstyle\pm2.16}$          & $61.39{\,\scriptstyle\pm4.61}$          & $84.19{\,\scriptstyle\pm2.04}$ \\
\midrule
\multirow{4}{*}{\textit{German}}  & \texttt{Qwen3-8B-DE}            & {\setlength{\fboxsep}{2pt}\setlength{\fboxrule}{0.7pt}\fcolorbox{soberblue}{white}{\textsc{de}}} & \cellcolor{cyan!15}$72.59$          & $93.12{\,\scriptstyle\pm0.94}$          & $75.15{\,\scriptstyle\pm1.19}$          & $55.20{\,\scriptstyle\pm2.30}$          & $54.56{\,\scriptstyle\pm3.33}$          & $84.94{\,\scriptstyle\pm1.12}$ \\
                                  & \texttt{Qwen3-8B-DE-Swap}       & {\setlength{\fboxsep}{2pt}\setlength{\fboxrule}{0.7pt}\fcolorbox{soberblue}{white}{\textsc{de}}} & \cellcolor{cyan!15}$75.15$          & $\mathbf{96.96}{\,\scriptstyle\pm0.89}$ & $77.35{\,\scriptstyle\pm1.24}$          & $56.16{\,\scriptstyle\pm2.49}$          & $58.28{\,\scriptstyle\pm4.12}$          & $\mathbf{87.00}{\,\scriptstyle\pm1.24}$ \\
                                  & \texttt{Qwen3-8B-DE-Pivot-EN}  & \textsc{en} & \cellcolor{cyan!15}$\mathbf{75.67}$ & $93.76{\,\scriptstyle\pm0.93}$          & $\mathbf{78.05}{\,\scriptstyle\pm0.98}$ & $\mathbf{57.68}{\,\scriptstyle\pm2.64}$ & $\mathbf{62.06}{\,\scriptstyle\pm3.81}$ & $86.81{\,\scriptstyle\pm1.63}$ \\
                                  & \texttt{Qwen3-8B-EN}            & \textsc{en} & \cellcolor{cyan!15}$73.53$          & $95.88{\,\scriptstyle\pm0.89}$          & $75.80{\,\scriptstyle\pm1.29}$          & $55.45{\,\scriptstyle\pm2.97}$          & $57.94{\,\scriptstyle\pm4.46}$          & $82.56{\,\scriptstyle\pm1.49}$ \\
\midrule
\multirow{4}{*}{\textit{Spanish}} & \texttt{Qwen3-8B-ES}            & {\setlength{\fboxsep}{2pt}\setlength{\fboxrule}{0.7pt}\fcolorbox{soberblue}{white}{\textsc{es}}} & \cellcolor{cyan!15}$72.41$          & $93.20{\,\scriptstyle\pm0.98}$          & $76.58{\,\scriptstyle\pm1.37}$          & $55.15{\,\scriptstyle\pm2.54}$          & $56.11{\,\scriptstyle\pm4.18}$          & $81.00{\,\scriptstyle\pm2.59}$ \\
                                  & \texttt{Qwen3-8B-ES-Swap}       & {\setlength{\fboxsep}{2pt}\setlength{\fboxrule}{0.7pt}\fcolorbox{soberblue}{white}{\textsc{es}}} & \cellcolor{cyan!15}$\mathbf{74.80}$ & $\mathbf{97.08}{\,\scriptstyle\pm1.05}$ & $77.10{\,\scriptstyle\pm0.94}$          & $55.15{\,\scriptstyle\pm2.87}$          & $58.50{\,\scriptstyle\pm4.11}$          & $\mathbf{86.19}{\,\scriptstyle\pm1.40}$ \\
                                  & \texttt{Qwen3-8B-ES-Pivot-EN}  & \textsc{en} & \cellcolor{cyan!15}$74.32$          & $95.08{\,\scriptstyle\pm0.98}$          & $78.20{\,\scriptstyle\pm1.24}$          & $\mathbf{56.57}{\,\scriptstyle\pm2.33}$ & $\mathbf{61.33}{\,\scriptstyle\pm4.43}$ & $80.44{\,\scriptstyle\pm1.84}$ \\
                                  & \texttt{Qwen3-8B-EN}            & \textsc{en} & \cellcolor{cyan!15}$74.41$          & $94.76{\,\scriptstyle\pm0.93}$          & $\mathbf{78.55}{\,\scriptstyle\pm1.52}$ & $54.44{\,\scriptstyle\pm2.14}$          & $61.06{\,\scriptstyle\pm4.32}$          & $83.25{\,\scriptstyle\pm2.02}$ \\
\midrule
\multirow{4}{*}{\textit{Chinese}} & \texttt{Qwen3-8B-ZH}            & {\setlength{\fboxsep}{2pt}\setlength{\fboxrule}{0.7pt}\fcolorbox{soberblue}{white}{\textsc{zh}}} & \cellcolor{cyan!15}$70.80$          & $88.92{\,\scriptstyle\pm2.13}$          & $74.85{\,\scriptstyle\pm1.46}$          & $50.71{\,\scriptstyle\pm2.29}$          & $53.89{\,\scriptstyle\pm4.30}$          & $85.62{\,\scriptstyle\pm1.59}$ \\
                                  & \texttt{Qwen3-8B-ZH-Swap}       & {\setlength{\fboxsep}{2pt}\setlength{\fboxrule}{0.7pt}\fcolorbox{soberblue}{white}{\textsc{zh}}} & \cellcolor{cyan!15}$71.62$          & $88.24{\,\scriptstyle\pm2.22}$          & $\mathbf{76.42}{\,\scriptstyle\pm1.25}$ & $52.58{\,\scriptstyle\pm3.14}$          & $55.17{\,\scriptstyle\pm3.77}$          & $\mathbf{85.69}{\,\scriptstyle\pm1.92}$ \\
                                  & \texttt{Qwen3-8B-ZH-Pivot-EN}  & \textsc{en} & \cellcolor{cyan!15}$\mathbf{73.89}$ & $\mathbf{94.84}{\,\scriptstyle\pm1.80}$ & $76.15{\,\scriptstyle\pm1.49}$          & $\mathbf{54.19}{\,\scriptstyle\pm2.44}$ & $\mathbf{59.06}{\,\scriptstyle\pm3.91}$ & $85.19{\,\scriptstyle\pm0.93}$ \\
                                  & \texttt{Qwen3-8B-EN}            & \textsc{en} & \cellcolor{cyan!15}$66.49$          & $76.04{\,\scriptstyle\pm2.55}$          & $75.00{\,\scriptstyle\pm1.01}$          & $47.53{\,\scriptstyle\pm2.97}$          & $50.00{\,\scriptstyle\pm4.89}$          & $83.88{\,\scriptstyle\pm1.93}$ \\
\midrule
\multirow{4}{*}{\textit{Swahili}} & \texttt{Qwen3-8B-SW}            & {\setlength{\fboxsep}{2pt}\setlength{\fboxrule}{0.7pt}\fcolorbox{soberblue}{white}{\textsc{sw}}} & \cellcolor{cyan!15}$66.98$          & $93.16{\,\scriptstyle\pm1.55}$          & $61.98{\,\scriptstyle\pm2.23}$          & $49.39{\,\scriptstyle\pm1.83}$          & $47.67{\,\scriptstyle\pm3.35}$          & $82.69{\,\scriptstyle\pm1.02}$ \\
                                  & \texttt{Qwen3-8B-SW-Swap}       & {\setlength{\fboxsep}{2pt}\setlength{\fboxrule}{0.7pt}\fcolorbox{soberblue}{white}{\textsc{sw}}} & \cellcolor{cyan!15}$69.09$          & $\mathbf{96.12}{\,\scriptstyle\pm0.98}$ & $64.10{\,\scriptstyle\pm2.16}$          & $49.29{\,\scriptstyle\pm1.74}$          & $50.33{\,\scriptstyle\pm3.59}$          & $\mathbf{85.62}{\,\scriptstyle\pm2.10}$ \\
                                  & \texttt{Qwen3-8B-SW-Pivot-EN}  & \textsc{en} & \cellcolor{cyan!15}$\mathbf{70.52}$ & $89.68{\,\scriptstyle\pm0.62}$          & $\mathbf{66.00}{\,\scriptstyle\pm2.26}$ & $\mathbf{52.73}{\,\scriptstyle\pm3.31}$ & $\mathbf{59.67}{\,\scriptstyle\pm4.00}$ & $84.50{\,\scriptstyle\pm1.28}$ \\
                                  & \texttt{Qwen3-8B-EN}            & \textsc{en} & \cellcolor{cyan!15}$37.96$          & $35.88{\,\scriptstyle\pm1.68}$          & $33.88{\,\scriptstyle\pm1.72}$          & $36.82{\,\scriptstyle\pm3.13}$          & $24.78{\,\scriptstyle\pm3.44}$          & $58.44{\,\scriptstyle\pm1.36}$ \\
\bottomrule
\end{tabular*}
\caption{Detailed per-language evaluation, for \textsc{xx} $\in$ \{\textsc{fr}, \textsc{de}, \textsc{es}, \textsc{zh}, \textsc{sw}\}: \texttt{Qwen3-8B-XX} (native reasoning in \textsc{xx}), \texttt{Qwen3-8B-XX-Pivot-EN} (English-pivoted reasoning, same \textsc{xx} Q\&A pairs as \texttt{Qwen3-8B-XX}), \texttt{Qwen3-8B-XX-Swap} (Layer Swap from \texttt{Qwen3-8B-EN} into \texttt{Qwen3-8B-XX}; see \S\ref{sec:layer-swap}), and \texttt{Qwen3-8B-EN} (English specialist, reference). Bold marks the best score per column within each language group. $\pm$ denotes the sample standard deviation across runs.}
\label{tab:main-results}
\end{table*}

\begin{figure*}[t]
    \centering
    \includegraphics[width=0.95\textwidth]{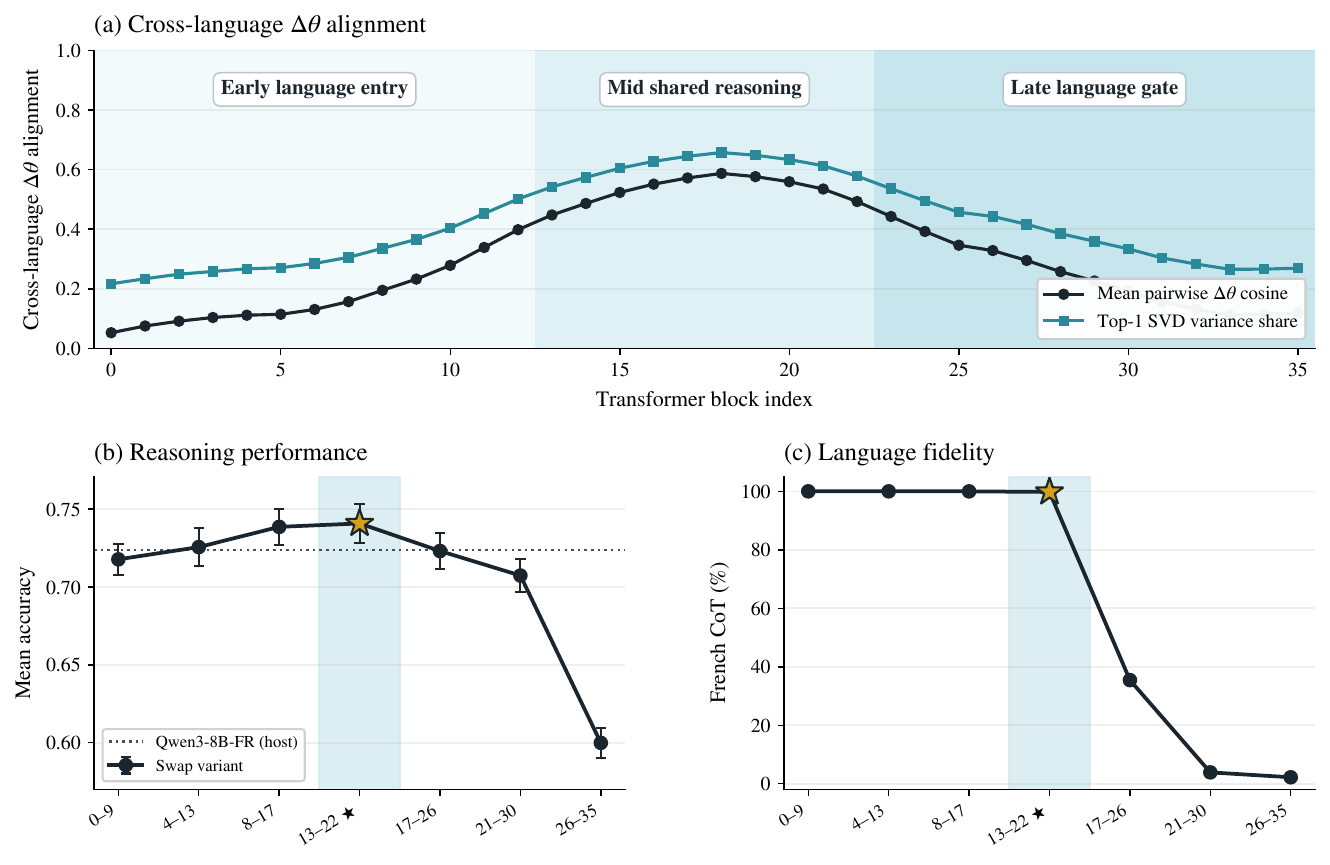}
    \caption{\textbf{(a)} Cross-language alignment of the per-language SFT updates $\Delta\theta_L^{(i)}$ as a function of transformer layer index $L$: mean pairwise cosine $\bar{c}_L$ and top-1 SVD variance share $s_L$ across the six per-language specialists. \textbf{(b)} Mean accuracy on the five French benchmarks for the seven Swap variants, each transferring a contiguous 10-layer window from \texttt{Qwen3-8B-EN} into \texttt{Qwen3-8B-FR}. \textbf{(c)} Language fidelity, the fraction of generated reasoning traces classified as French, for each Swap variant.}
    \label{fig:layer-swap-ablation}
\end{figure*}

\subsection{Matched-Supervision Experiments}
\label{sec:native-vs-pivoted}

To compare native and English-pivoted reasoning under matched supervision, we train, for each non-English language \textsc{xx} $\in \{\textsc{fr}, \textsc{de}, \textsc{es}, \textsc{zh}, \textsc{sw}\}$, two specialists from \texttt{Qwen/Qwen3-8B-Base} on the same per-language question--answer pairs, differing only in the CoT language: a native specialist \texttt{Qwen3-8B-XX} whose reasoning trace is also in \textsc{xx}, and an English-pivoted specialist \texttt{Qwen3-8B-XX-Pivot-EN} whose questions and answers remain in \textsc{xx} but whose reasoning trace is generated in English. Both are trained with comparable per-language token budgets of $10$--$11$B and identical hyperparameters (Table~\ref{tab:training-data-composition}, Appendix). Note that sample counts differ slightly because samples exceeding the 32K-token training context are dropped, which disproportionately affects non-English versions due to less efficient tokenization (e.g. Swahili) or higher verbosity (e.g. French, Spanish). We also include the English specialist \texttt{Qwen3-8B-EN} for reference, but the controlled comparison remains \texttt{Qwen3-8B-XX} vs.\ \texttt{Qwen3-8B-XX-Pivot-EN}: comparing \texttt{Qwen3-8B-XX} directly to \texttt{Qwen3-8B-EN} would conflate the CoT language with the fact that \texttt{Qwen3-8B-EN} has seen no \textsc{xx}-language supervision.

Under this setting, the native reasoning gap is substantially smaller than prior reports suggest. On the five-benchmark average, \texttt{Qwen3-8B-XX} trails \texttt{Qwen3-8B-XX-Pivot-EN} by $2.7\%$, $3.1\%$, $1.9\%$, $3.1\%$, and $3.5\%$ on French, German, Spanish, Chinese, and Swahili respectively (Table~\ref{tab:main-results}). The gap broadly tracks the resource tier: smallest on the high-resource languages ($1.9$--$3.1\%$) and largest on low-resource Swahili ($3.5\%$). Per benchmark, AIME 24/25 dominates the gap, with native specialists trailing their English-pivoted counterparts by $5$--$7\%$ on French, German, Spanish, and Chinese and by $12\%$ on Swahili; the other benchmarks contribute much smaller differences (around $1$--$4\%$ on average), and HumanEvalPlus is essentially tied across languages. The lower-resource end is also where native SFT contributes the most in absolute terms: \texttt{Qwen3-8B-EN} scores only $37.96\%$ on the Swahili five-benchmark average, and the matched native specialist nearly doubles this to $66.98\%$, whereas the corresponding gain over \texttt{Qwen3-8B-EN} is marginal for the high-resource Latin-script trio.

\section{Layer Swap}
\label{sec:layer-swap}

\subsection{Method and Layer Selection}

\textbf{Layer Swap}~\citep{bandarkar2024layer, bandarkar2025unreasonable} is a training-free model-composition technique~\citep{ilharco2023editing}. Starting from a shared base model, two \emph{experts} are independently fine-tuned and recomposed into a single hybrid model by replacing a contiguous range of transformer layers in one expert with the corresponding layers from the other, while the remaining layers are kept unchanged. \citet{bandarkar2024layer} apply this to short-form math instruction tuning, combining a math-knowledge expert trained in English with a target-language fluency expert trained on generic instructions. To our knowledge, the technique has not been applied to long-CoT reasoning models, nor to a pair of experts that share the same reasoning skills but differ in their training language, the setting we study here. The two experts are the native specialist (e.g. \texttt{Qwen3-8B-FR}) and the English specialist (\texttt{Qwen3-8B-EN}); we recompose them by transferring a contiguous range of the English specialist's transformer layers into the native specialist.

\paragraph{Weight-space analysis}
We measure how the $N$ per-language specialists' weight updates $\Delta\theta_L^{(\textsc{xx})} = \theta_L^{(\textsc{xx})} - \theta_{\mathrm{base}}$ (for language $\textsc{xx} \in \{\textsc{en}, \textsc{fr}, \textsc{de}, \textsc{es}, \textsc{zh}, \textsc{sw}\}$ at layer $L$) agree across languages, layer by layer, using two complementary statistics; in our setup $N = 6$ (English, French, German, Spanish, Chinese, and Swahili). The first is the mean pairwise cosine of the $N$ deltas,
\begin{equation}
\bar{c}_L = \binom{N}{2}^{-1} \!\!\sum_{\textsc{xx} < \textsc{yy}} \frac{\langle \Delta\theta_L^{(\textsc{xx})},\, \Delta\theta_L^{(\textsc{yy})} \rangle}{\|\Delta\theta_L^{(\textsc{xx})}\|\,\|\Delta\theta_L^{(\textsc{yy})}\|},
\end{equation}
which captures local alignment between pairs of language-specific deltas. The second is the top-1 SVD variance share: stacking the $N$ deltas as rows of a matrix $D_L \in \mathbb{R}^{N \times P_L}$ (where $P_L$ is the number of parameters in layer $L$) whose singular values are $\sigma_1 \ge \dots \ge \sigma_N$,
\begin{equation}
s_L = \frac{\sigma_1^2}{\sum_{k=1}^{N}\sigma_k^2},
\end{equation}
capturing the fraction of cross-language variance concentrated along a single direction in weight space. Both statistics are low at the two ends of the stack, rise sharply between L9 and L13, and reach a high range across L13--L22, where $\bar{c}_L \approx 0.6$ and $s_L$ captures $65$--$74\%$ of the cross-language variance (Figure~\ref{fig:layer-swap-ablation}; (a)), with per-layer delta norms remaining comparable across the stack (Appendix~\ref{sec:appendix}, Figure~\ref{fig:fig_delta_norm}). In this window, the six per-language SFT updates align along a shared cross-language direction (a largely language-agnostic component of the SFT update) while at the early and late ends of the stack they diverge along language-specific directions. This depth structure is consistent with prior weight-space analyses of multilingual layer specialization~\citep{bandarkar2024layer, tang2024language}; we verify it here for long-CoT native reasoning specialists across six languages.

\paragraph{Motivation}
This mid-stack cross-lingual alignment has a natural interpretation. Current multilingual LLMs appear to route reasoning through English-aligned intermediate representations~\citep{wendler2024llamas, schut2025multilingual, etxaniz2024multilingual, zhao2024how}. In an English specialist, the middle layers may therefore develop reasoning circuits directly on top of representations already aligned with the pretraining distribution, which is not the case for a native specialist. This asymmetry may limit how effectively the native specialist's central layers specialize for reasoning during post-training, yielding a stronger reasoning core in the English specialist's middle layers. Transferring only the English middle layers into the native specialist should therefore install part of the English reasoning advantage while leaving the native edge layers, and with them the target-language CoT, intact. We test whether this swap improves reasoning while preserving target-language CoT generation.

\paragraph{Layer-range ablation}
We conduct the layer-range ablation on the French specialist as a case study. Transferring layers from the English specialist (\texttt{Qwen3-8B-EN}) into the French specialist (\texttt{Qwen3-8B-FR}), we sweep the transferred layer range from L0 to L35 in contiguous windows and report two quantities per configuration: average accuracy on the French benchmarks (Figure~\ref{fig:layer-swap-ablation}; (b)), and \emph{language fidelity} (Figure~\ref{fig:layer-swap-ablation}; (c)), the fraction of generated reasoning traces classified as French by a FastText language identifier~\citep{grave2018learning}. Fidelity stays at $\sim$100\% for any window confined to L0--L22 and collapses for windows extending further, placing the CoT-language gate around L22. Accuracy improves materially only when the English specialist's mid-stack is included; early-only swaps preserve fidelity but transfer little of its reasoning capability. The L13--L22 window satisfies both criteria, preserving native CoT generation while capturing the English specialist's reasoning advantage. Replacing the English source specialist with either the Chinese or the German specialist eliminates the improvement (Table~\ref{tab:swap-source-ablation}, Appendix~\ref{sec:swap-source-ablation}), confirming that the effect originates specifically from the English specialist's middle-layer representations.

\subsection{Results}

In Table~\ref{tab:main-results}, we compare \texttt{Qwen3-8B-XX-Swap}, obtained by transplanting a mid-stack window from the English specialist into the native specialist, against the three models introduced in \S\ref{sec:native-vs-pivoted}. For French, German, Spanish, and Swahili, we use the L13--L22 window identified above. For Chinese, however, this window causes roughly $60\%$ of traces to revert to English mid-generation, suggesting that the target-language gate is located slightly earlier in the stack. We therefore use L13--L20 instead, which fully restores Chinese CoT fidelity (Appendix~\ref{sec:chinese-swap-ablation}).

In particular, the swapped models retain the target reasoning language while closing most of the native reasoning gap relative to English pivoting. Measured as the fraction of the native-vs-pivoted gap closed on the five-benchmark average, Swap closes $89\%$ of the gap on French, $83\%$ on German, $27\%$ on Chinese, $60\%$ on Swahili, and matches the English-pivoted ceiling on Spanish. In absolute terms, the remaining gap to \texttt{Qwen3-8B-XX-Pivot-EN} shrinks to between $0.0\%$ and $2.3\%$ across languages. The gap is essentially closed for the Latin-script European trio, while the remaining differences on Chinese and Swahili reflect, respectively, typological distance from English and lower-resource status.

Across benchmarks, Layer Swap leads on HumanEvalPlus in all five languages and on MGSM-Rev2 in most of them, while the remaining gap concentrates on the more reasoning-intensive tasks (Global-MMLU-Lite, GPQA-Diamond, and AIME). On AIME 24/25 specifically, Layer Swap narrows the gap to just $\sim$3\% on French, German, Spanish, and Chinese, with Swahili remaining the outlier at $\sim$9\%. Language fidelity remains at $\sim$100\% across all five Swap variants. We also test whether transplanting the English specialist’s mid-layers introduces English-centric bias on culturally grounded content using \emph{Global-PIQA}~\citep{chang2025global}, a commonsense benchmark written natively by native-speaker researchers and containing many culturally specific items. Swap matches or slightly outperforms native specialists across all five languages while preserving target-language CoT (Appendix~\ref{sec:global-piqa}). These results suggest that mid-layer transplantation does not introduce noticeable English-centric contamination.

\section{The Understanding Gap}

As an additional ablation enabled by our per-language specialists, we isolate the input-understanding component of the multilingual reasoning gap, a question \citet{kang2025multilingual} address with inference-time interventions on \texttt{Qwen/Qwen3-4B} but that remains open under large-scale native SFT. We evaluate each specialist on the native and English versions of every benchmark; each model continues to reason in its training language in both conditions without any prompting or constraint, so only the input language varies.

Every non-English specialist scores higher on English inputs than on native inputs despite never seeing English in SFT (Table~\ref{tab:crosslingual-average}; per-benchmark numbers in Table~\ref{tab:crosslingual-detailed}, Appendix). The improvement grows sharply with the language's distance from English in the base model, a $\sim$4$\times$ effect on Swahili relative to the European trio. Two effects drive this. \textbf{(i)} Pretraining data imbalance: the Qwen3 base most likely saw considerably more English and Latin-script European tokens than Swahili during pretraining (the exact mixture is not disclosed), and our 10B-token SFT only retargets the reasoning policy on top of the input-side representations inherited from that pretraining. \textbf{(ii)} Intermediate representations in English-centric base models are more strongly aligned with the English representation space for high-resource Latin-script languages than for typologically distant or low-resource ones~\citep{wendler2024llamas, schut2025multilingual}, so the representational gap between input-side embeddings and the English-aligned reasoning mid-stack is largest precisely for Swahili.

\begin{table}[h]
\centering
\setlength{\tabcolsep}{4pt}
\renewcommand{\arraystretch}{1.08}
\rowcolors{3}{gray!10}{white}

\begin{tabular}{lccc}
\toprule
\rowcolor{white}
\multirow{2}{*}{\textbf{Model}}
& \multicolumn{2}{c}{\textbf{Evaluation language}}
& \multirow{2}{*}{$\boldsymbol{\Delta}$} \\
\cmidrule(lr){2-3}
\rowcolor{white}
& \textbf{English} & \textbf{Native} & \\
\midrule
\texttt{Qwen3-8B-EN} & \textbf{77.00} & --    & -- \\
\texttt{Qwen3-8B-ES} & \textbf{73.72} & 72.41 & $-1.31$ \\
\texttt{Qwen3-8B-DE} & \textbf{73.94} & 72.59 & $-1.35$ \\
\texttt{Qwen3-8B-FR} & \textbf{73.78} & 72.36 & $-1.42$ \\
\texttt{Qwen3-8B-ZH} & \textbf{74.54} & 70.80 & $-3.74$ \\
\texttt{Qwen3-8B-SW} & \textbf{72.43} & 66.98 & $-5.45$ \\
\bottomrule
\end{tabular}
\caption{Average accuracy of each per-language specialist, averaged over MGSM-Rev2, Global-MMLU-Lite, GPQA-Diamond, AIME 24/25, and HumanEvalPlus, on the English and native versions of each benchmark.}
\label{tab:crosslingual-average}
\rowcolors{2}{}{}
\end{table}

\section{Conclusion}

We revisited the native reasoning gap in multilingual LLMs under matched supervision. We trained native and English-pivoted specialists from \texttt{Qwen/Qwen3-8B-Base} on the same per-language Q\&A pairs using $\sim$10B-token budgets and 32k context windows across French, German, Spanish, Chinese, and Swahili. Across five evaluation benchmarks, the performance gap shrank to $1.9$--$3.5\%$ on average, substantially smaller than prior reports suggested, with the remaining differences concentrated in complex reasoning-intensive mathematics tasks. A weight-space analysis revealed a largely language-agnostic reasoning core in the mid-stack, motivating a training-free Layer Swap that transferred these English layers into each native specialist while preserving $\sim$100\% language fidelity. On the five-benchmark average, the gap was fully closed on Spanish, reduced by $83$--$89$\% on French and German, and by $27$--$60$\% on Chinese and Swahili. Since English-pivoted reasoning is the default in essentially all open reasoning models, the required English specialist is available off the shelf, making the recipe directly applicable to other model families.

\section{Limitations}

While this work establishes a controlled study of the native reasoning gap, several limitations remain. First, our main analysis was limited to a single model family and parameter scale. As an initial cross-architecture probe, we replicated the French pipeline on \texttt{HuggingFaceTB/SmolLM3-3B-Base}, and observed a native-vs-pivoted gap and a Layer Swap behaviour consistent with what we report on \texttt{Qwen/Qwen3-8B-Base} (Appendix~\ref{sec:smollm3-3B-ablation}). Extending the study to a wider range of architectures, scales, and languages would nonetheless strengthen the findings. The recipe also presupposed a multilingually pretrained base and covered only six languages, with Swahili as the sole lower-resource representative. In addition, tokenization efficiency varied across languages, meaning that an equivalent token budget may have corresponded to less effective supervision for languages such as Swahili. Our primary control was a matched token-budget comparison rather than a matched-example comparison. A stricter design would train both regimes on the intersection of examples retained in both native and English-pivoted form; we leave this paired control to future work.

Second, we did not perform an extensive hyperparameter sweep for our $\sim$10B-token SFT runs, nor did we exhaustively explore the Layer Swap window, and finer-grained sweeps over its position and width could identify a better configuration. While our training budget was substantial for native long-CoT supervision, it remained below production-scale post-training settings, where larger token budgets and multi-stage curricula may have led to different outcomes.

Finally, as is common in multilingual long-CoT post-training, our corpus is machine-translated, so the reported native-vs.-English-pivoted gap should be interpreted within this translated supervision setting. However, because translation noise primarily affects the native specialists, stronger translation models or higher-quality data would likely narrow the gap further; the reported $1.9$--$3.5\%$ range should therefore be viewed as an upper bound. Our evaluation focused on mathematics, science, general knowledge, and code generation; extending it to tasks with stronger cultural or sociopragmatic components would provide a broader assessment of multilingual reasoning. Moreover, our specialist models were trained solely with SFT, and combining the Swap models with RL or preference tuning remains a natural direction for future work.

\section{Ethical Considerations}

In terms of broader impact, our specialists maintain CoT in the input language across six languages, including the lower-resource Swahili, making reasoning models more interpretable for non-English users and reducing reliance on English as a mandatory intermediate language. On the risk side, the released specialists inherit biases from both \texttt{Qwen/Qwen3-8B-Base} and the translation model used to construct our corpus. They have not undergone additional safety tuning, and may therefore exhibit unsafe or undesirable behaviors outside the scope of our evaluation setting.

\section{Acknowledgments}

We thank Adrien Cavaillès, Théo Lasnier, and Armel Randy Zebaze for helpful comments and discussions. We also gratefully acknowledge the EuroHPC Joint Undertaking for awarding us Fast Lane allocations on MareNostrum 5, hosted by the Barcelona Supercomputing Center, under applications EHPC-AIF-2025FL01-571 and EHPC-AIF-2026FL01-212, which provided the computational resources used in this work. This project is also supported by the OpenEuroLLM project, co-funded by the Digital Europe Programme under GA no. 101195233. For more information see \url{https://openeurollm.eu}.


\bibliography{custom}

@article{ghosh2025survey,
  title={A survey of multilingual reasoning in language models},
  author={Ghosh, Akash and Datta, Debayan and Saha, Sriparna and Agarwal, Chirag},
  journal={arXiv preprint arXiv:2502.09457},
  year={2025}
}

@article{peter2025mind,
  title={Mind the Gap... or Not? How Translation Errors and Evaluation Details Skew Multilingual Results},
  author={Peter, Jan-Thorsten and Vilar, David and Domhan, Tobias and Malkin, Dan and Freitag, Markus},
  journal={arXiv preprint arXiv:2511.05162},
  year={2025}
}

@misc{singh2024globalmmluunderstandingaddressing,
      title={Global MMLU: Understanding and Addressing Cultural and Linguistic Biases in Multilingual Evaluation}, 
      author={Shivalika Singh and Angelika Romanou and Clémentine Fourrier and David I. Adelani and Jian Gang Ngui and Daniel Vila-Suero and Peerat Limkonchotiwat and Kelly Marchisio and Wei Qi Leong and Yosephine Susanto and Raymond Ng and Shayne Longpre and Wei-Yin Ko and Madeline Smith and Antoine Bosselut and Alice Oh and Andre F. T. Martins and Leshem Choshen and Daphne Ippolito and Enzo Ferrante and Marzieh Fadaee and Beyza Ermis and Sara Hooker},
      year={2024},
      eprint={2412.03304},
      archivePrefix={arXiv},
      primaryClass={cs.CL},
      url={https://arxiv.org/abs/2412.03304}, 
}

@misc{olmo2025olmo3,
title={Olmo 3},
author={Team Olmo and Allyson Ettinger and Amanda Bertsch and Bailey Kuehl and David Graham and David Heineman and Dirk Groeneveld and Faeze Brahman and Finbarr Timbers and Hamish Ivison and Jacob Morrison and Jake Poznanski and Kyle Lo and Luca Soldaini and Matt Jordan and Mayee Chen and Michael Noukhovitch and Nathan Lambert and Pete Walsh and Pradeep Dasigi and Robert Berry and Saumya Malik and Saurabh Shah and Scott Geng and Shane Arora and Shashank Gupta and Taira Anderson and Teng Xiao and Tyler Murray and Tyler Romero and Victoria Graf and Akari Asai and Akshita Bhagia and Alexander Wettig and Alisa Liu and Aman Rangapur and Chloe Anastasiades and Costa Huang and Dustin Schwenk and Harsh Trivedi and Ian Magnusson and Jaron Lochner and Jiacheng Liu and Lester James V. Miranda and Maarten Sap and Malia Morgan and Michael Schmitz and Michal Guerquin and Michael Wilson and Regan Huff and Ronan Le Bras and Rui Xin and Rulin Shao and Sam Skjonsberg and Shannon Zejiang Shen and Shuyue Stella Li and Tucker Wilde and Valentina Pyatkin and Will Merrill and Yapei Chang and Yuling Gu and Zhiyuan Zeng and Ashish Sabharwal and Luke Zettlemoyer and Pang Wei Koh and Ali Farhadi and Noah A. Smith and Hannaneh Hajishirzi},
year={2025},
eprint={2512.13961},
archivePrefix={arXiv},
primaryClass={cs.CL},
url={https://arxiv.org/abs/2512.13961},
}

@article{shi2022language,
  title={Language models are multilingual chain-of-thought reasoners},
  author={Shi, Freda and Suzgun, Mirac and Freitag, Markus and Wang, Xuezhi and Srivats, Suraj and Vosoughi, Soroush and Chung, Hyung Won and Tay, Yi and Ruder, Sebastian and Zhou, Denny and others},
  journal={arXiv preprint arXiv:2210.03057},
  year={2022}
}

@article{yang2025qwen3,
  title={Qwen3 technical report},
  author={Yang, An and Li, Anfeng and Yang, Baosong and Zhang, Beichen and Hui, Binyuan and Zheng, Bo and Yu, Bowen and Gao, Chang and Huang, Chengen and Lv, Chenxu and others},
  journal={arXiv preprint arXiv:2505.09388},
  year={2025}
}

@article{guo2025deepseek,
  title={Deepseek-r1: Incentivizing reasoning capability in llms via reinforcement learning},
  author={Guo, Daya and Yang, Dejian and Zhang, Haowei and Song, Junxiao and Wang, Peiyi and Zhu, Qihao and Xu, Runxin and Zhang, Ruoyu and Ma, Shirong and Bi, Xiao and others},
  journal={arXiv preprint arXiv:2501.12948},
  year={2025}
}

@article{rein2023gpqa,
  title={Gpqa: A graduate-level google-proof q\&a benchmark},
  author={Rein, David and Hou, Betty Li and Stickland, Asa Cooper and Petty, Jackson and Pang, Richard Yuanzhe and Dirani, Julien and Michael, Julian and Bowman, Samuel R},
  journal={arXiv preprint arXiv:2311.12022},
  year={2023}
}

@article{shah2024flashattention,
  title={Flashattention-3: Fast and accurate attention with asynchrony and low-precision},
  author={Shah, Jay and Bikshandi, Ganesh and Zhang, Ying and Thakkar, Vijay and Ramani, Pradeep and Dao, Tri},
  journal={Advances in Neural Information Processing Systems},
  volume={37},
  pages={68658--68685},
  year={2024}
}

@article{jacobs2023deepspeed,
  title={Deepspeed ulysses: System optimizations for enabling training of extreme long sequence transformer models},
  author={Jacobs, Sam Ade and Tanaka, Masahiro and Zhang, Chengming and Zhang, Minjia and Song, Shuaiwen Leon and Rajbhandari, Samyam and He, Yuxiong},
  journal={arXiv preprint arXiv:2309.14509},
  year={2023}
}

@article{liu2023your,
  title={Is your code generated by chatgpt really correct? rigorous evaluation of large language models for code generation},
  author={Liu, Jiawei and Xia, Chunqiu Steven and Wang, Yuyao and Zhang, Lingming},
  journal={Advances in neural information processing systems},
  volume={36},
  pages={21558--21572},
  year={2023}
}

@article{son2025pushing,
  title={Pushing on Multilingual Reasoning Models with Language-Mixed Chain-of-Thought},
  author={Son, Guijin and Yang, Donghun and Patel, Hitesh Laxmichand and Agarwal, Amit and Ko, Hyunwoo and Lim, Chanuk and Panda, Srikant and Kim, Minhyuk and Drolia, Nikunj and Choi, Dasol and others},
  journal={arXiv preprint arXiv:2510.04230},
  year={2025}
}

@misc{vonwerra2020trl,
  title   = {{TRL: Transformers Reinforcement Learning}},
  author  = {von Werra, Leandro and Belkada, Younes and Tunstall, Lewis and Beeching, Edward and Thrush, Tristan and Lambert, Nathan and Huang, Shengyi and Rasul, Kashif and Gallou\'edec, Quentin},
  howpublished = {\url{https://github.com/huggingface/trl}},
  year    = {2020}
}

@inproceedings{bandarkar2025unreasonable,
  title={The Unreasonable Effectiveness of Model Merging for Cross-Lingual Transfer in LLMs},
  author={Bandarkar, Lucas and Peng, Nanyun},
  booktitle={Proceedings of the 5th Workshop on Multilingual Representation Learning (MRL 2025)},
  pages={131--148},
  year={2025}
}

@article{bandarkar2024layer,
  title={Layer swapping for zero-shot cross-lingual transfer in large language models},
  author={Bandarkar, Lucas and Muller, Benjamin and Yuvraj, Pritish and Hou, Rui and Singhal, Nayan and Lv, Hongjiang and Liu, Bing},
  journal={arXiv preprint arXiv:2410.01335},
  year={2024}
}

@article{park2025cross,
  title={Cross-lingual collapse: How language-centric foundation models shape reasoning in large language models},
  author={Park, Cheonbok and Kim, Jeonghoon and Lee, Joosung and Bae, Sanghwan and Choo, Jaegul and Yoo, Kang Min},
  journal={arXiv preprint arXiv:2506.05850},
  year={2025}
}

@article{huang2025beyond,
  title={Beyond English-Centric Training: How Reinforcement Learning Improves Cross-Lingual Reasoning in LLMs},
  author={Huang, Shulin and Ding, Yiran and Pan, Junshu and Zhang, Yue},
  journal={arXiv preprint arXiv:2509.23657},
  year={2025}
}

@article{zhang2025think,
  title={Think Natively: Unlocking Multilingual Reasoning with Consistency-Enhanced Reinforcement Learning},
  author={Zhang, Xue and Liang, Yunlong and Meng, Fandong and Zhang, Songming and Huang, Kaiyu and Chen, Yufeng and Xu, Jinan and Zhou, Jie},
  journal={arXiv preprint arXiv:2510.07300},
  year={2025}
}

@article{shao2024deepseekmath,
  title={Deepseekmath: Pushing the limits of mathematical reasoning in open language models},
  author={Shao, Zhihong and Wang, Peiyi and Zhu, Qihao and Xu, Runxin and Song, Junxiao and Bi, Xiao and Zhang, Haowei and Zhang, Mingchuan and Li, YK and Wu, Yang and others},
  journal={arXiv preprint arXiv:2402.03300},
  year={2024}
}

@article{gurgurov2026reasonxl,
  title={ReasonXL: Shifting LLM Reasoning Language Without Sacrificing Performance},
  author={Gurgurov, Daniil and R{\"o}hr, Tom and von Rohrscheidt, Sebastian and van Genabith, Josef and L{\"o}ser, Alexander and Ostermann, Simon},
  journal={arXiv preprint arXiv:2604.12378},
  year={2026}
}

@misc{qi2025modelsreasonlanguagecontrolling,
      title={When Models Reason in Your Language: Controlling Thinking Trace Language Comes at the Cost of Accuracy}, 
      author={Jirui Qi and Shan Chen and Zidi Xiong and Raquel Fernández and Danielle S. Bitterman and Arianna Bisazza},
      year={2025},
      eprint={2505.22888},
      archivePrefix={arXiv},
      primaryClass={cs.CL},
      url={https://arxiv.org/abs/2505.22888}, 
}

@misc{eval-harness,
  author       = {Gao, Leo and Tow, Jonathan and Abbasi, Baber and Biderman, Stella and Black, Sid and DiPofi, Anthony and Foster, Charles and Golding, Laurence and Hsu, Jeffrey and Le Noac'h, Alain and Li, Haonan and McDonell, Kyle and Muennighoff, Niklas and Ociepa, Chris and Phang, Jason and Reynolds, Laria and Schoelkopf, Hailey and Skowron, Aviya and Sutawika, Lintang and Tang, Eric and Thite, Anish and Wang, Ben and Wang, Kevin and Zou, Andy},
  title        = {The Language Model Evaluation Harness},
  month        = 07,
  year         = 2024,
  publisher    = {Zenodo},
  version      = {v0.4.3},
  doi          = {10.5281/zenodo.12608602},
  url          = {https://zenodo.org/records/12608602}
}

@article{barua2025long,
  title={Long chain-of-thought reasoning across languages},
  author={Barua, Josh and Eisape, Seun and Yin, Kayo and Suhr, Alane},
  journal={arXiv preprint arXiv:2508.14828},
  year={2025}
}

@inproceedings{saji2026reasoning,
  title={The reasoning lingua franca: A double-edged sword for multilingual AI},
  author={Saji, Alan and Dabre, Raj and Kunchukuttan, Anoop and Puduppully, Ratish},
  booktitle={Proceedings of the 19th Conference of the European Chapter of the Association for Computational Linguistics (Volume 2: Short Papers)},
  pages={329--344},
  year={2026}
}

@inproceedings{wendler2024llamas,
  title={Do llamas work in english? on the latent language of multilingual transformers},
  author={Wendler, Chris and Veselovsky, Veniamin and Monea, Giovanni and West, Robert},
  booktitle={Proceedings of the 62nd Annual Meeting of the Association for Computational Linguistics (Volume 1: Long Papers)},
  pages={15366--15394},
  year={2024}
}

@article{schut2025multilingual,
  title={Do multilingual llms think in english?},
  author={Schut, Lisa and Gal, Yarin and Farquhar, Sebastian},
  journal={arXiv preprint arXiv:2502.15603},
  year={2025}
}

@inproceedings{etxaniz2024multilingual,
  title={Do multilingual language models think better in English?},
  author={Etxaniz, Julen and Azkune, Gorka and Soroa, Aitor and de Lacalle, Oier Lopez and Artetxe, Mikel},
  booktitle={Proceedings of the 2024 Conference of the North American Chapter of the Association for Computational Linguistics: Human Language Technologies (Volume 2: Short Papers)},
  pages={550--564},
  year={2024}
}

@article{kang2025multilingual,
  title={Why Do Multilingual Reasoning Gaps Emerge in Reasoning Language Models?},
  author={Kang, Deokhyung and Hwang, Seonjeong and Kim, Daehui and Kim, Hyounghun and Lee, Gary Geunbae},
  journal={arXiv preprint arXiv:2510.27269},
  year={2025}
}

@inproceedings{grave2018learning,
  title={Learning Word Vectors for 157 Languages},
  author={Grave, Edouard and Bojanowski, Piotr and Gupta, Prakhar and Joulin, Armand and Mikolov, Tomas},
  booktitle={Proceedings of the International Conference on Language Resources and Evaluation (LREC 2018)},
  year={2018}
}

@inproceedings{tang2024language,
  title={Language-Specific Neurons: The Key to Multilingual Capabilities in Large Language Models},
  author={Tang, Tianyi and Luo, Wenyang and Huang, Haoyang and Zhang, Dongdong and Wang, Xiaolei and Zhao, Xin and Wei, Furu and Wen, Ji-Rong},
  booktitle={Proceedings of the 62nd Annual Meeting of the Association for Computational Linguistics (Volume 1: Long Papers)},
  pages={5701--5715},
  year={2024}
}

@inproceedings{rajbhandari2020zero,
  title={Zero: Memory optimizations toward training trillion parameter models},
  author={Rajbhandari, Samyam and Rasley, Jeff and Ruwase, Olatunji and He, Yuxiong},
  booktitle={SC20: international conference for high performance computing, networking, storage and analysis},
  pages={1--16},
  year={2020},
  organization={IEEE}
}

@inproceedings{zhao2024how,
  title={How do Large Language Models Handle Multilingualism?},
  author={Zhao, Yiran and Zhang, Wenxuan and Chen, Guizhen and Kawaguchi, Kenji and Bing, Lidong},
  booktitle={Advances in Neural Information Processing Systems},
  year={2024}
}

@inproceedings{ilharco2023editing,
  title={Editing Models with Task Arithmetic},
  author={Ilharco, Gabriel and Ribeiro, Marco Tulio and Wortsman, Mitchell and Gururangan, Suchin and Schmidt, Ludwig and Hajishirzi, Hannaneh and Farhadi, Ali},
  booktitle={The Eleventh International Conference on Learning Representations},
  year={2023}
}

@article{jaech2024openai,
  title={Openai o1 system card},
  author={Jaech, Aaron and Kalai, Adam and Lerer, Adam and Richardson, Adam and El-Kishky, Ahmed and Low, Aiden and Helyar, Alec and Madry, Aleksander and Beutel, Alex and Carney, Alex and others},
  journal={arXiv preprint arXiv:2412.16720},
  year={2024}
}

@inproceedings{Gailly2012zlibCL,
  title={zlib compression library},
  author={Jean-Loup Gailly and M Adler},
  year={2012},
  url={https://api.semanticscholar.org/CorpusID:60948258}
}

@article{chang2025global,
  title={Global PIQA: Evaluating Commonsense Reasoning Across 100+ Languages and Cultures},
  author={Chang, Tyler A and Arnett, Catherine and Sadallah, Abdelrahman and Eldesokey, Abdelrahman and Kashar, Abeer and Daud, Abolade and Olanihun, Abosede Grace and Mohammed, Adamu Labaran and Praise, Adeyemi and Sharma, Adhikarimayum Meerajita and others},
  journal={arXiv preprint arXiv:2510.24081},
  year={2025}
}

\clearpage
\appendix
\section{Appendix}
\label{sec:appendix}

\raggedbottom
\renewcommand{\topfraction}{0.95}
\renewcommand{\bottomfraction}{0.95}
\renewcommand{\textfraction}{0.05}
\renewcommand{\floatpagefraction}{0.80}
\renewcommand{\dbltopfraction}{0.95}
\renewcommand{\dblfloatpagefraction}{0.80}
\setcounter{topnumber}{4}
\setcounter{bottomnumber}{4}
\setcounter{totalnumber}{8}
\setcounter{dbltopnumber}{4}

\FloatBarrier 

\subsection{Dataset Statistics}

Table~\ref{tab:dataset_stats} reports the category mix of the English source corpus, dominated by mathematics and code with long mean trace lengths. Table~\ref{tab:dataset_language_stats} gives the per-language sample counts and total trained tokens after filtering using the Qwen3 tokenizer.

\begin{table}[H]
\centering
\setlength{\tabcolsep}{3pt}
\renewcommand{\arraystretch}{1.08}
\rowcolors{2}{gray!10}{white}
\begin{tabular}{lrrr}
\toprule
\rowcolor{white}
\textbf{Category} & \textbf{Count} & \textbf{\%} & \textbf{Mean Tok.} \\
\midrule
math           & 184,284 & 37.93 & 19,369 \\
code           & 169,102 & 34.80 &  7,340 \\
instr.-follow. &  69,543 & 14.31 &  2,310 \\
science        &  25,290 &  5.21 &  9,089 \\
safety         &  19,351 &  3.98 &    739 \\
general-chat   &  17,262 &  3.55 &  3,076 \\
struct.-data   &   1,041 &  0.21 &  2,228 \\
\bottomrule
\end{tabular}
\rowcolors{2}{}{}
\caption{Category distribution and mean token length of the English subset of \emph{Dolci-Think-SFT-32B}.}
\label{tab:dataset_stats}
\end{table}

\begin{table}[H]
\centering
\renewcommand{\arraystretch}{1.12}
\begin{tabular}{lrrr}
\toprule
\rowcolor{white}
\multirow{2}{*}{\textbf{Language}} &
\multirow{2}{*}{\textbf{Samples}} &
\multicolumn{2}{c}{\textbf{Tokens}} \\
\cmidrule(lr){3-4}
\rowcolor{white}
 & & \multicolumn{1}{c}{\textbf{Mean}} & \multicolumn{1}{c}{\textbf{Total}} \\
\midrule
\arrayrulecolor{white}
\rowcolor{white}
English & 485,873 & 10,848 & 5.27B \\ \hline
\rowcolor{gray!6}
Chinese & 484,078 & 10,890 & 5.27B \\ \hline
\rowcolor{gray!10}
Spanish & 479,748 & 13,400 & 6.43B \\ \hline
\rowcolor{gray!14}
French  & 483,588 & 13,702 & 6.63B \\ \hline
\rowcolor{gray!18}
German   & 480,529 & 13,846 & 6.65B \\ \hline
\rowcolor{gray!22}
Swahili & 473,582 & 16,208 & 7.68B \\
\arrayrulecolor{black}
\bottomrule
\end{tabular}
\caption{Per-language statistics of the native reasoning SFT dataset: number of samples, mean sample length, and total token count.}
\label{tab:dataset_language_stats}
\end{table}

\subsection{Translation}

Table~\ref{tab:translation_prompt_de} gives the chunk-level translation prompt applied independently to question, reasoning trace, and final answer at $\sim$2k-token boundaries. We perform translation with \texttt{google/gemma-3-27b-it} at a temperature of $0.15$. For reference, translating 1B tokens with \texttt{google/gemma-3-27b-it} required approximately $750$ H100 GPU-hours in our setup.

\begin{table}[!tbp]
\centering
\small
\setlength{\tabcolsep}{5pt}
\renewcommand{\arraystretch}{1.08}

\begin{tabular}{|p{0.94\columnwidth}|}
\hline
\cellcolor{gray!10}
\begin{minipage}{0.90\columnwidth}
\vspace{0.6em}
\ttfamily\footnotesize\raggedright
You are a translation engine. Your ONLY task is to translate the text inside <source> into <LANGUAGE>.

\vspace{0.45em}
HARD RULES (must follow):\\
- Output ONLY the translation, and nothing else.\\
- Do NOT answer the message or continue the conversation.\\
- Do NOT add advice, role labels, extra lines, or formatting.\\
- Translate EVERYTHING inside <source> completely.\\
- Use informal address unless formality or plurality is clearly required.\\
- Preserve meaning, intent, tone, and punctuation as naturally as possible.

\vspace{0.45em}
SPECIAL CASES:\\
- For code, do NOT translate identifiers. Translate only comments and human-language strings, preserving whitespace and indentation.\\
- For units, convert to metric while keeping calculations consistent.

\vspace{0.45em}
INPUT:\\
<source>\\
\{message\_text\}\\
</source>

\vspace{0.45em}
OUTPUT: <LANGUAGE> translation only.
\vspace{0.6em}
\end{minipage}
\\
\hline
\end{tabular}

\caption{Prompt template used for our translation pipeline.}
\label{tab:translation_prompt_de}
\end{table}

\subsection{Evaluation Prompts}
\label{sec:eval_prompts}

We list the language-specific instruction templates passed to \texttt{lm-eval-harness} for each benchmark family. Prompts were translated and adapted from the English originals while preserving the answer-extraction format used by the harness. Table~\ref{tab:prompts_aime} covers AIME 24/25, Table~\ref{tab:prompts_mgsm} MGSM-Rev2, Table~\ref{tab:prompts_global_mmlu} Global-MMLU-Lite, Table~\ref{tab:prompts_gpqa} GPQA-Diamond, and Table~\ref{tab:prompts_humanevalplus} HumanEvalPlus.

\begin{table}[!tbp]
\centering
\footnotesize
\setlength{\tabcolsep}{6pt}
\renewcommand{\arraystretch}{1.30}
\rowcolors{2}{gray!10}{white}
\begin{tabular}{|c|p{0.78\columnwidth}|}
\hline
\rowcolor{white}
\textbf{Lang} & \textbf{Suffix appended after \texttt{\{\{problem\}\}\textbackslash n}} \\
\hline
en & Please reason step by step, and put your final answer within \texttt{\textbackslash boxed\{\}}. \\
de & Bitte denke Schritt für Schritt nach und gib deine endgültige Antwort in \texttt{\textbackslash boxed\{\}} an. \\
fr & Veuillez raisonner étape par étape, et mettre votre réponse finale dans \texttt{\textbackslash boxed\{\}}. \\
es & Por favor, razona paso a paso y pon tu respuesta final en \texttt{\textbackslash boxed\{\}}. \\
zh & \begin{CJK}{UTF8}{gbsn}请一步步推理，并将最终答案放在\end{CJK} \texttt{\textbackslash boxed\{\}} \begin{CJK}{UTF8}{gbsn}内。\end{CJK} \\
sw & Tafadhali fikiri hatua kwa hatua, na uweke jibu lako la mwisho ndani ya \texttt{\textbackslash boxed\{\}}. \\
\hline
\end{tabular}
\caption{Prompt template used to evaluate the multilingual AIME24 and AIME25 benchmarks in \texttt{lm-eval-harness}.}
\label{tab:prompts_aime}
\end{table}

\begin{table}[!tbp]
\centering
\footnotesize
\setlength{\tabcolsep}{6pt}
\renewcommand{\arraystretch}{1.30}
\rowcolors{2}{gray!10}{white}
\begin{tabular}{|c|p{0.78\columnwidth}|}
\hline
\rowcolor{white}
\textbf{Lang} & \textbf{Prefix preceding \texttt{\textbackslash n\{\{question\}\}}} \\
\hline
en & Solve this math problem. Give the reasoning steps before giving the final answer on the last line by itself in the format of ``Answer:''. Do not add anything other than the integer answer after ``Answer:''. \\
de & Löse dieses Mathematikproblem. Gib die Schritte zur Begründung an, bevor du die endgültige Antwort in der letzten Zeile alleine im Format ``Antwort:'' gibst. Füge nichts anderes als die ganzzahlige Antwort nach ``Antwort:'' hinzu. \\
fr & Résolvez ce problème de mathématiques. Donnez les étapes de raisonnement avant de fournir la réponse finale sur la dernière ligne elle-même dans le format de ``Réponse:''. N'ajoutez rien d'autre que le nombre de la réponse après ``Réponse:''. \\
es & Resuelve este problema matemático. Proporciona los pasos de razonamiento antes de dar la respuesta final en la última línea por sí misma en el formato de ``Respuesta:''. No añadas nada más que la respuesta entera después de ``Respuesta:''. \\
zh & \begin{CJK}{UTF8}{gbsn}解决这个数学问题。在最后一行给出答案前，请提供推理步骤。最后一行应该以``答案：''的形式独立给出答案。在``答案：''后不要添加除整数答案之外的任何内容。\end{CJK} \\
sw & Suluhisha tatizo hili la hesabu. Toa hatua za mantiki kabla ya kutoa jibu la mwisho kwenye mstari wa mwisho peke yake katika muundo wa ``Jibu:''. Usiongeze chochote kingine isipokuwa jibu la integer baada ya ``Jibu:''. \\
\hline
\end{tabular}
\caption{Prompt template used to evaluate the MGSM-Rev2 benchmark in \texttt{lm-eval-harness}.}
\label{tab:prompts_mgsm}
\end{table}

\begin{table}[!tbp]
\centering
\footnotesize
\setlength{\tabcolsep}{6pt}
\renewcommand{\arraystretch}{1.30}
\rowcolors{2}{gray!10}{white}
\begin{tabular}{|c|p{0.78\columnwidth}|}
\hline
\rowcolor{white}
\textbf{Lang} & \textbf{Prefix preceding the question and four options} \\
\hline
en & Answer the following multiple choice question. On the last line of your response, place your final answer letter within \texttt{\textbackslash boxed\{\}} (e.g. \texttt{\textbackslash boxed\{A\}}), ensuring only the letter is inside. Think step by step before answering. \\
de & Beantworten Sie die folgende Multiple-Choice-Frage. Setzen Sie in der letzten Zeile Ihrer Antwort den Buchstaben Ihrer endgültigen Antwort in \texttt{\textbackslash boxed\{\}} (z.B. \texttt{\textbackslash boxed\{A\}}). Denken Sie Schritt für Schritt, bevor Sie antworten. \\
fr & Répondez à la question suivante à choix multiples. Sur la dernière ligne de votre réponse, placez la lettre de votre réponse finale dans \texttt{\textbackslash boxed\{\}} (par exemple, \texttt{\textbackslash boxed\{A\}}). Réfléchissez étape par étape avant de répondre. \\
es & Responda la siguiente pregunta de opción múltiple. En la última línea de su respuesta, coloque la letra de su respuesta final dentro de \texttt{\textbackslash boxed\{\}} (por ejemplo, \texttt{\textbackslash boxed\{A\}}). Piense paso a paso antes de responder. \\
zh & \begin{CJK}{UTF8}{gbsn}回答以下多项选择题。在回答的最后一行，将你的最终答案字母放在\end{CJK} \texttt{\textbackslash boxed\{\}} \begin{CJK}{UTF8}{gbsn}中（例如\end{CJK} \texttt{\textbackslash boxed\{A\}}\begin{CJK}{UTF8}{gbsn}），确保框内只有字母。请在回答前逐步思考。\end{CJK} \\
sw & Jibu swali lifuatalo la chaguo nyingi. Kwenye mstari wa mwisho wa jibu lako, weka herufi ya jibu lako la mwisho ndani ya \texttt{\textbackslash boxed\{\}} (kwa mfano, \texttt{\textbackslash boxed\{A\}}), hakikisha herufi pekee iko ndani. Fikiria hatua kwa hatua kabla ya kujibu. \\
\hline
\end{tabular}
\caption{Prompt prefix used to evaluate the Global-MMLU benchmark in \texttt{lm-eval-harness}; the question and four options A/B/C/D follow on subsequent lines.}
\label{tab:prompts_global_mmlu}
\end{table}

\begin{table}[!tbp]
\centering
\footnotesize
\setlength{\tabcolsep}{6pt}
\renewcommand{\arraystretch}{1.30}
\rowcolors{2}{gray!10}{white}
\begin{tabular}{|c|p{0.78\columnwidth}|}
\hline
\rowcolor{white}
\textbf{Lang} & \textbf{Suffix appended after \texttt{\{\{problem\}\}\textbackslash n\textbackslash n}} \\
\hline
en & Think step by step before answering. \\
de & Denken Sie Schritt für Schritt nach, bevor Sie antworten. \\
fr & Réfléchissez étape par étape avant de répondre. \\
es & Piensa paso a paso antes de responder. \\
zh & \begin{CJK}{UTF8}{gbsn}请一步步思考后再回答。\end{CJK} \\
sw & Fikiria hatua kwa hatua kabla ya kujibu. \\
\hline
\end{tabular}
\caption{Prompt template used to evaluate the GPQA-Diamond benchmark in \texttt{lm-eval-harness}. The dataset's \texttt{problem} field already contains the question and four options A/B/C/D.}
\label{tab:prompts_gpqa}
\end{table}

\begin{table}[!tbp]
\centering
\footnotesize
\setlength{\tabcolsep}{6pt}
\renewcommand{\arraystretch}{1.30}
\rowcolors{2}{gray!10}{white}
\begin{tabular}{|c|p{0.78\columnwidth}|}
\hline
\rowcolor{white}
\textbf{Lang} & \textbf{\texttt{doc\_to\_text} (\texttt{\{\{ prompt \}\}} is the function signature + docstring)} \\
\hline
en & Write a solution to the following problem and make sure that it passes the tests:\textbackslash n\verb!```python!\textbackslash n\texttt{\{\{ prompt \}\}}\textbackslash n\verb!```!\textbackslash n Please reason step by step, and put your final answer within a \verb!```python! code layer.\textbackslash n \\
de & Schreiben Sie eine Lösung für das folgende Problem und stellen Sie sicher, dass sie die Tests besteht:\textbackslash n\verb!```python!\textbackslash n\texttt{\{\{ prompt \}\}}\textbackslash n\verb!```!\textbackslash n Bitte denken Sie Schritt für Schritt und geben Sie Ihre endgültige Antwort in einem \verb!```python! Codelayer an.\textbackslash n \\
fr & Écrivez une solution au problème suivant et assurez-vous qu'elle passe les tests :\textbackslash n\verb!```python!\textbackslash n\texttt{\{\{ prompt \}\}}\textbackslash n\verb!```!\textbackslash n Veuillez raisonner étape par étape, et mettre votre réponse finale dans un bloc de code \verb!```python!.\textbackslash n \\
es & Escribe una solución al siguiente problema y asegúrate de que pase las pruebas:\textbackslash n\verb!```python!\textbackslash n\texttt{\{\{ prompt \}\}}\textbackslash n\verb!```!\textbackslash n Por favor, razona paso a paso y coloca tu respuesta final dentro de un bloque de código \verb!```python!.\textbackslash n \\
zh & \begin{CJK}{UTF8}{gbsn}编写以下问题的解决方案，并确保它通过测试：\end{CJK}\textbackslash n\verb!```python!\textbackslash n\texttt{\{\{ prompt \}\}}\textbackslash n\verb!```!\textbackslash n\begin{CJK}{UTF8}{gbsn}请逐步推理，并将最终答案放在\end{CJK} \verb!```python! \begin{CJK}{UTF8}{gbsn}代码块中。\end{CJK}\textbackslash n \\
sw & Andika suluhisho la tatizo lifuatalo na uhakikishe kuwa linapita majaribio:\textbackslash n\verb!```python!\textbackslash n\texttt{\{\{ prompt \}\}}\textbackslash n\verb!```!\textbackslash n Tafadhali fikiria hatua kwa hatua, na uweke jibu lako la mwisho ndani ya layer ya msimbo \verb!```python!.\textbackslash n \\
\hline
\end{tabular}
\caption{Prompt template used to evaluate the HumanEvalPlus benchmark in \texttt{lm-eval-harness}.}
\label{tab:prompts_humanevalplus}
\end{table}

\subsection{Training Settings}

Table~\ref{tab:sft_key_hparams} lists the SFT hyperparameters used to train every per-language specialist, held identical across the native and English-pivoted regimes.

\begin{table}[!tbp]
\centering
\small
\setlength{\tabcolsep}{5pt}
\renewcommand{\arraystretch}{1.08}
\rowcolors{2}{gray!10}{white}

\begin{tabular}{|p{0.46\columnwidth}|p{0.42\columnwidth}|}
\hline
\rowcolor{white}
\textbf{Hyperparameter} & \textbf{Setting} \\
\hline
Base model              & Qwen3-8B-Base \\
Training parallelism    & ZeRO-3 \& Ulysses SP \\
Attention backend       & FlashAttention 3 \\
Maximum sequence length & 32,768 tokens \\
Global batch size       & $\sim$1M tokens \\
Optimizer               & AdamW \\
Learning rate           & $5 \times 10^{-5}$ \\
Warmup ratio            & 0.1 \\
Weight decay            & 0.01 \\
Training epochs         & 2 \\
Numerical precision     & bf16 \\
Gradient checkpointing  & Enabled \\
Sequence packing        & Enabled \\
Loss masking            & Assistant-only \\
Liger kernels           & Enabled \\
\hline
\end{tabular}

\caption{SFT hyperparameters, identical across all per-language specialists and across the native and English-pivoted regimes.}
\label{tab:sft_key_hparams}
\end{table}

\subsection{Detailed Training Results}

Table~\ref{tab:training-data-composition} reports the unique sample count and total training-token count for each specialist. Table~\ref{tab:scaling-scores} expands the scaling curves of Figure~\ref{fig:scaling_law_avg} with per-budget mean accuracies in every language. Table~\ref{tab:crosslingual-detailed} extends the cross-lingual evaluation of Table~\ref{tab:crosslingual-average} with per-benchmark scores on both the English and native versions of every benchmark.


\begin{table*}[!htbp]
\centering
\footnotesize
\setlength{\tabcolsep}{4pt}
\renewcommand{\arraystretch}{1.05}
\begin{tabular}{@{}llcrr@{}}
\toprule
\textbf{Lang} & \textbf{Model} & \textbf{CoT} & \textbf{Unique Samples} & \textbf{Trained Tokens} \\
\midrule
\textit{English}                  & \texttt{Qwen3-8B-EN}            & \textsc{en} & 480,625 & 10.40B \\
\midrule
\multirow{2}{*}{\textit{French}}  & \texttt{Qwen3-8B-FR}            & \textsc{fr} & 447,650 & 10.92B \\
                                  & \texttt{Qwen3-8B-FR-Pivot-EN}  & \textsc{en} & 478,255 & 10.53B \\
\midrule
\multirow{2}{*}{\textit{German}}  & \texttt{Qwen3-8B-DE}            & \textsc{de} & 443,100 & 10.85B \\
                                  & \texttt{Qwen3-8B-DE-Pivot-EN}  & \textsc{en} & 475,205 & 10.54B \\
\midrule
\multirow{2}{*}{\textit{Spanish}} & \texttt{Qwen3-8B-ES}            & \textsc{es} & 448,984 & 10.90B \\
                                  & \texttt{Qwen3-8B-ES-Pivot-EN}  & \textsc{en} & 474,463 & 10.48B \\
\midrule
\multirow{2}{*}{\textit{Chinese}} & \texttt{Qwen3-8B-ZH}            & \textsc{zh} & 478,622 & 10.39B \\
                                  & \texttt{Qwen3-8B-ZH-Pivot-EN}  & \textsc{en} & 478,811 & 10.39B \\
\midrule
\multirow{2}{*}{\textit{Swahili}} & \texttt{Qwen3-8B-SW}            & \textsc{sw} & 400,844 & 10.03B \\
                                  & \texttt{Qwen3-8B-SW-Pivot-EN}  & \textsc{en} & 468,213 & 10.49B \\
\bottomrule
\end{tabular}
\caption{Training-data composition per specialist: number of unique samples and total trained tokens, after filtering and the 32K-token context cap.}
\label{tab:training-data-composition}
\end{table*}


\begin{table*}[!htbp]
\centering
\footnotesize
\setlength{\tabcolsep}{6pt}
\renewcommand{\arraystretch}{1.05}
\begin{tabular*}{\textwidth}{@{\extracolsep{\fill}}lccccc@{}}
\toprule
\textbf{Language} & $\sim$\textbf{100M tokens} & $\sim$\textbf{400M tokens} & $\sim$\textbf{1B tokens} & $\sim$\textbf{2B tokens} & $\sim$\textbf{10B tokens} \\
\midrule
English & $62.62{\,\scriptstyle\pm1.31}$ & $66.47{\,\scriptstyle\pm1.52}$ & $68.42{\,\scriptstyle\pm1.54}$ & $69.83{\,\scriptstyle\pm1.23}$ & $\mathbf{71.51}{\,\scriptstyle\pm1.31}$ \\
French  & $57.48{\,\scriptstyle\pm1.55}$ & $61.24{\,\scriptstyle\pm1.40}$ & $63.70{\,\scriptstyle\pm1.47}$ & $64.74{\,\scriptstyle\pm1.40}$ & $67.26{\,\scriptstyle\pm1.35}$ \\
German  & $55.83{\,\scriptstyle\pm1.14}$ & $61.51{\,\scriptstyle\pm1.51}$ & $62.59{\,\scriptstyle\pm1.55}$ & $64.59{\,\scriptstyle\pm1.38}$ & $67.46{\,\scriptstyle\pm1.09}$ \\
Spanish & $57.52{\,\scriptstyle\pm1.29}$ & $62.16{\,\scriptstyle\pm1.30}$ & $63.70{\,\scriptstyle\pm1.53}$ & $64.61{\,\scriptstyle\pm1.28}$ & $67.21{\,\scriptstyle\pm1.43}$ \\
Chinese & $55.46{\,\scriptstyle\pm1.43}$ & $60.09{\,\scriptstyle\pm1.36}$ & $61.95{\,\scriptstyle\pm1.50}$ & $63.54{\,\scriptstyle\pm1.36}$ & $66.27{\,\scriptstyle\pm1.33}$ \\
Swahili & $38.67{\,\scriptstyle\pm1.14}$ & $48.56{\,\scriptstyle\pm1.52}$ & $53.82{\,\scriptstyle\pm1.27}$ & $56.47{\,\scriptstyle\pm1.42}$ & $60.43{\,\scriptstyle\pm1.13}$ \\
\bottomrule
\end{tabular*}
\caption{Mean accuracy of the six per-language specialists at each SFT-token budget, averaged across MGSM-Rev2, Global-MMLU-Lite, GPQA-Diamond, AIME 24/25, and HumanEvalPlus in the training language. Token counts are approximate. $\pm$ denotes the sample standard deviation across runs.}
\label{tab:scaling-scores}
\end{table*}


\begin{table*}[!htbp]
\centering
\footnotesize
\setlength{\tabcolsep}{4pt}
\renewcommand{\arraystretch}{1.05}
\begin{tabular*}{\textwidth}{@{\extracolsep{\fill}}llccccccc@{}}
\toprule
\textbf{Model} & \textbf{Eval Lang} & \textbf{CoT} & \textbf{Avg} & \textbf{MGSM} & \textbf{G-MMLU} & \textbf{GPQA-D} & \textbf{AIME 24/25} & \textbf{HEval+} \\
\midrule
\texttt{Qwen3-8B-EN}                  & \textsc{en} & \textsc{en} & $77.00$ & $98.96{\,\scriptstyle\pm0.47}$          & $81.72{\,\scriptstyle\pm1.30}$          & $55.66{\,\scriptstyle\pm2.05}$          & $62.89{\,\scriptstyle\pm4.22}$          & $85.75{\,\scriptstyle\pm1.88}$ \\
\midrule
\multirow{2}{*}{\texttt{Qwen3-8B-FR}} & \textsc{fr} & \textsc{fr} & $72.36$          & $92.80{\,\scriptstyle\pm1.71}$          & $76.45{\,\scriptstyle\pm1.73}$          & $53.59{\,\scriptstyle\pm2.85}$          & $55.67{\,\scriptstyle\pm3.93}$          & $83.31{\,\scriptstyle\pm1.56}$ \\
                                      & \textsc{en} & \textsc{fr} & $\mathbf{73.78}$ & $\mathbf{94.20}{\,\scriptstyle\pm1.53}$ & $\mathbf{79.30}{\,\scriptstyle\pm1.15}$ & $\mathbf{55.25}{\,\scriptstyle\pm2.58}$ & $\mathbf{56.61}{\,\scriptstyle\pm4.02}$ & $\mathbf{83.56}{\,\scriptstyle\pm2.06}$ \\
\midrule
\multirow{2}{*}{\texttt{Qwen3-8B-DE}} & \textsc{de} & \textsc{de} & $72.59$          & $93.12{\,\scriptstyle\pm0.94}$          & $75.15{\,\scriptstyle\pm1.19}$          & $\mathbf{55.20}{\,\scriptstyle\pm2.30}$ & $54.56{\,\scriptstyle\pm3.33}$          & $84.94{\,\scriptstyle\pm1.12}$ \\
                                      & \textsc{en} & \textsc{de} & $\mathbf{73.94}$ & $\mathbf{94.76}{\,\scriptstyle\pm1.12}$ & $\mathbf{79.83}{\,\scriptstyle\pm1.22}$ & $54.04{\,\scriptstyle\pm2.14}$          & $\mathbf{55.94}{\,\scriptstyle\pm3.91}$ & $\mathbf{85.13}{\,\scriptstyle\pm1.79}$ \\
\midrule
\multirow{2}{*}{\texttt{Qwen3-8B-ES}} & \textsc{es} & \textsc{es} & $72.41$          & $93.20{\,\scriptstyle\pm0.98}$          & $76.58{\,\scriptstyle\pm1.37}$          & $\mathbf{55.15}{\,\scriptstyle\pm2.54}$ & $56.11{\,\scriptstyle\pm4.18}$          & $81.00{\,\scriptstyle\pm2.59}$ \\
                                      & \textsc{en} & \textsc{es} & $\mathbf{73.72}$ & $\mathbf{94.48}{\,\scriptstyle\pm1.45}$ & $\mathbf{79.90}{\,\scriptstyle\pm1.39}$ & $\mathbf{55.15}{\,\scriptstyle\pm2.28}$ & $\mathbf{56.67}{\,\scriptstyle\pm3.47}$ & $\mathbf{82.38}{\,\scriptstyle\pm1.21}$ \\
\midrule
\multirow{2}{*}{\texttt{Qwen3-8B-ZH}} & \textsc{zh} & \textsc{zh} & $70.80$          & $88.92{\,\scriptstyle\pm2.13}$          & $74.85{\,\scriptstyle\pm1.46}$          & $50.71{\,\scriptstyle\pm2.29}$          & $53.89{\,\scriptstyle\pm4.30}$          & $\mathbf{85.62}{\,\scriptstyle\pm1.59}$ \\
                                      & \textsc{en} & \textsc{zh} & $\mathbf{74.54}$ & $\mathbf{95.96}{\,\scriptstyle\pm1.34}$ & $\mathbf{79.45}{\,\scriptstyle\pm1.39}$ & $\mathbf{54.14}{\,\scriptstyle\pm2.14}$ & $\mathbf{58.61}{\,\scriptstyle\pm3.69}$ & $84.56{\,\scriptstyle\pm0.84}$ \\
\midrule
\multirow{2}{*}{\texttt{Qwen3-8B-SW}} & \textsc{sw} & \textsc{sw} & $66.98$          & $93.16{\,\scriptstyle\pm1.55}$          & $61.98{\,\scriptstyle\pm2.23}$          & $49.39{\,\scriptstyle\pm1.83}$          & $47.67{\,\scriptstyle\pm3.35}$          & $82.69{\,\scriptstyle\pm1.02}$ \\
                                      & \textsc{en} & \textsc{sw} & $\mathbf{72.43}$ & $\mathbf{95.04}{\,\scriptstyle\pm1.00}$ & $\mathbf{78.98}{\,\scriptstyle\pm1.30}$ & $\mathbf{51.87}{\,\scriptstyle\pm2.08}$ & $\mathbf{51.56}{\,\scriptstyle\pm3.81}$ & $\mathbf{84.69}{\,\scriptstyle\pm1.45}$ \\
\bottomrule
\end{tabular*}
\caption{Detailed evaluation of each per-language specialist on the English and native versions of MGSM-Rev2, Global-MMLU-Lite, GPQA-Diamond, AIME 24/25, and HumanEvalPlus. Bold marks the best score per column within each model. $\pm$ denotes the sample standard deviation across runs.}
\label{tab:crosslingual-detailed}
\end{table*}

\subsection{Global-PIQA Cultural Evaluation}
\label{sec:global-piqa}
As a check that Layer Swap does not introduce English-centric bias on culturally grounded content, we evaluate on \emph{Global-PIQA}~\citep{chang2025global}, a commonsense benchmark natively authored by over 350 native-speaker researchers, with more than half of its items referencing local customs and culture. Table~\ref{tab:global-piqa} shows that Swap matches or exceeds the native specialist across all five languages, indicating no English-centric contamination from the transplanted mid-layers.

\begin{table*}[!tp]
\centering
\footnotesize
\setlength{\tabcolsep}{6pt}
\renewcommand{\arraystretch}{1.1}
\begin{tabular*}{\textwidth}{@{\extracolsep{\fill}}lccccc@{}}
\toprule
\textbf{Variant} & \textbf{French} & \textbf{German} & \textbf{Spanish} & \textbf{Chinese} & \textbf{Swahili} \\
\midrule
\texttt{Qwen3-8B-XX}          & $\mathbf{77.6}$ & $74.2$          & $83.0$          & $\mathbf{75.0}$ & $80.8$ \\
\texttt{Qwen3-8B-XX-Pivot-EN} & $77.5$          & $76.0$          & $\mathbf{85.1}$ & $74.2$          & $81.5$ \\
\texttt{Qwen3-8B-XX-Swap}     & $76.9$          & $\mathbf{76.3}$ & $84.7$          & $74.8$          & $\mathbf{83.1}$ \\
\bottomrule
\end{tabular*}
\caption{Global-PIQA accuracy for each per-language specialist and its Pivot-EN and Layer-Swap variants, averaged over 10 runs. Each column reports accuracy in the language named at the top. Bold marks the best variant per language.}
\label{tab:global-piqa}
\end{table*}

\subsection{SmolLM3-3B French Replication}
\label{sec:smollm3-3B-ablation}

The same pattern replicates on \texttt{SmolLM3-3B} (Table~\ref{tab:smollm3-fr-variants}): Pivot-EN yields the strongest French average, and Layer Swap closes most of the gap while leading on MGSM by preserving the base model's French arithmetic reasoning.

\begin{table*}[!tp]
\centering
\footnotesize
\setlength{\tabcolsep}{6pt}
\renewcommand{\arraystretch}{1.05}
\begin{tabular*}{\textwidth}{@{\extracolsep{\fill}}lcccccc@{}}
\toprule
\textbf{Model} & \textbf{Avg} & \textbf{MGSM} & \textbf{G-MMLU} & \textbf{GPQA-D} & \textbf{AIME 24/25} & \textbf{HEval+} \\
\midrule
\texttt{SmolLM3-3B-FR}          & $53.6$          & $84.4$          & $56.0$          & $30.9$          & $27.2$          & $69.4$ \\
\texttt{SmolLM3-3B-FR-Pivot-EN} & $\mathbf{55.5}$ & $82.8$          & $\mathbf{59.7}$ & $\mathbf{32.5}$ & $\mathbf{31.7}$ & $\mathbf{70.9}$ \\
\texttt{SmolLM3-3B-FR-Swap}     & $55.0$          & $\mathbf{89.0}$ & $58.9$          & $29.2$          & $28.0$          & $70.0$ \\
\bottomrule
\end{tabular*}
\caption{Comparison of \texttt{SmolLM3-3B-FR} variants across reasoning, knowledge, and code benchmarks. Bold marks the best score per column.}
\label{tab:smollm3-fr-variants}
\end{table*}

\subsection{Layer-Swap Source Language}
\label{sec:swap-source-ablation}

Table~\ref{tab:swap-source-ablation} verifies that the Layer Swap gain originates specifically from the English specialist's mid-layers: replacing \texttt{Qwen3-8B-EN} with the Chinese specialist \texttt{Qwen3-8B-ZH} as the swap source erases the improvement on French.

\begin{table*}[!htbp]
\centering
\footnotesize
\setlength{\tabcolsep}{4pt}
\renewcommand{\arraystretch}{1.05}
\begin{tabular*}{\textwidth}{@{\extracolsep{\fill}}llccccccc@{}}
\toprule
\textbf{Model} & \textbf{Source} & \textbf{Avg} & \textbf{MGSM} & \textbf{G-MMLU} & \textbf{GPQA-D} & \textbf{AIME 24/25} & \textbf{HEval+} \\
\midrule
\texttt{Qwen3-8B-FR-Swap}    & \textsc{en} & $\mathbf{74.74}$ & $\mathbf{97.40}{\,\scriptstyle\pm0.60}$ & $\mathbf{76.57}{\,\scriptstyle\pm1.54}$ & $\mathbf{54.55}{\,\scriptstyle\pm2.74}$ & $\mathbf{59.11}{\,\scriptstyle\pm4.08}$ & $\mathbf{86.06}{\,\scriptstyle\pm2.06}$ \\
\texttt{--} & \textsc{zh} & $72.34$          & $94.08{\,\scriptstyle\pm1.25}$          & $76.08{\,\scriptstyle\pm1.77}$          & $53.33{\,\scriptstyle\pm2.37}$          & $54.44{\,\scriptstyle\pm3.46}$          & $83.75{\,\scriptstyle\pm1.98}$ \\
\texttt{--} & \textsc{de} & $72.12$          & $92.24{\,\scriptstyle\pm1.25}$          & $76.08{\,\scriptstyle\pm1.36}$          & $52.78{\,\scriptstyle\pm2.47}$          & $56.11{\,\scriptstyle\pm4.09}$          & $83.38{\,\scriptstyle\pm2.13}$ \\
\bottomrule
\end{tabular*}
\caption{Layer-Swap source-language ablation on French: layers L13--L22 transferred into \texttt{Qwen3-8B-FR} from either the English specialist \texttt{Qwen3-8B-EN} (source \textsc{en}), the Chinese specialist \texttt{Qwen3-8B-ZH} (source \textsc{zh}), or the German specialist \texttt{Qwen3-8B-DE} (source \textsc{de}). Bold marks the best score per column. $\pm$ denotes the sample standard deviation across runs.}
\label{tab:swap-source-ablation}
\end{table*}

\subsection{Chinese Layer-Swap Window}
\label{sec:chinese-swap-ablation}

\begin{figure}[H]
    \centering
    \includegraphics[width=\columnwidth]{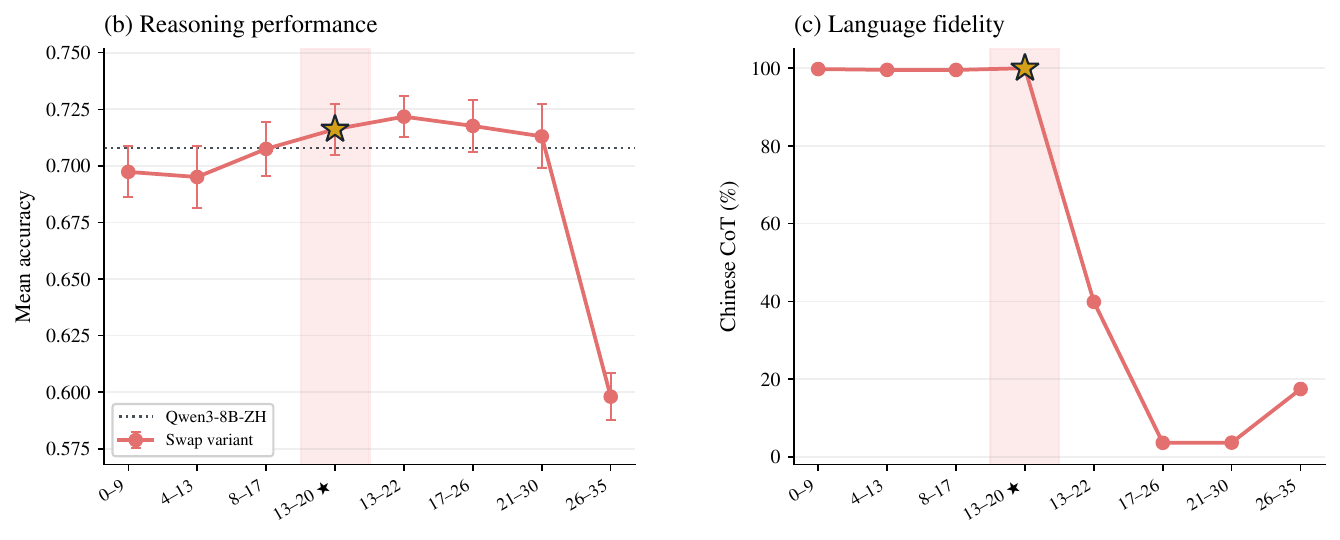}
    \caption{Same ablation as Figure~\ref{fig:layer-swap-ablation} (b,c) applied to \texttt{Qwen3-8B-ZH}: mean accuracy across the five Chinese benchmarks and Chinese language fidelity, as a function of the transferred layer window.}
    \label{fig:chinese-swap-ablation}
\end{figure}

Figure~\ref{fig:chinese-swap-ablation} repeats the layer-range ablation of Figure~\ref{fig:layer-swap-ablation} on \texttt{Qwen3-8B-ZH} and motivates the L13--L20 window adopted for \texttt{Qwen3-8B-ZH-Swap} since the L13--L22 window leaks $\sim$60\% of Chinese traces back into English.

\subsection{Per-Layer Update Magnitudes}

\begin{figure}[H]
    \centering
    \includegraphics[width=\columnwidth]{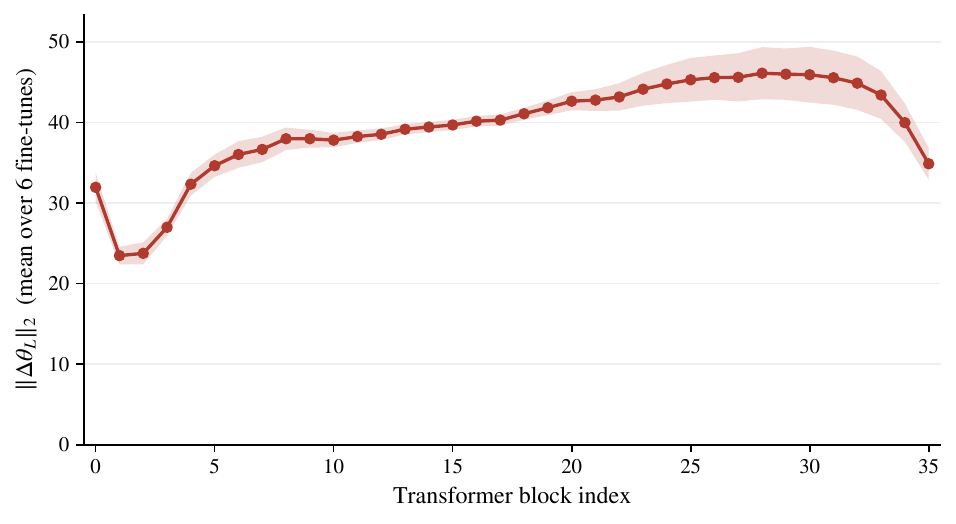}
    \caption{Per-layer L2 norm of the language-specific SFT update $\Delta\theta_L^{(\textsc{xx})} = \theta_L^{(\textsc{xx})} - \theta_{\mathrm{base}}$ across the six per-language specialists \textsc{xx} $\in$ \{\textsc{en}, \textsc{fr}, \textsc{de}, \textsc{es}, \textsc{zh}, \textsc{sw}\}; mean with $\pm$ std.}

    \label{fig:fig_delta_norm}
\end{figure}

Figure~\ref{fig:fig_delta_norm} reports the per-layer L2 norm of the per-language SFT updates. Norms remain comparable across the stack, confirming that the mid-stack agreement of Figure~\ref{fig:layer-swap-ablation} (a) reflects directional alignment rather than vanishing updates.

\end{document}